\documentclass[lettersize,journal]{IEEEtran}
\usepackage{gensymb}
\usepackage{amsmath,amsfonts}
\usepackage{amssymb}
\usepackage{bbm}
\usepackage{bm}
\usepackage{algorithmic}
\usepackage{algorithm}
\usepackage{array}
\usepackage[caption=false,font=scriptsize,labelfont=sf,textfont=sf]{subfig}
\usepackage{textcomp}
\usepackage{stfloats}
\usepackage{url}
\usepackage{verbatim}
\usepackage{graphicx}
\usepackage{booktabs}
\usepackage{tabularx}
\usepackage{multirow}
\usepackage{cite}
\usepackage{tikz}
\usepackage{multirow}
\newcolumntype{C}[1]{>{\centering\let\newline\\\arraybackslash\hspace{0pt}}m{#1}}
\newcolumntype{L}[1]{>{\raggedright\let\newline\\\arraybackslash\hspace{0pt}}m{#1}}
\newcolumntype{R}[1]{>{\raggedleft\let\newline\\\arraybackslash\hspace{0pt}}m{#1}}
\newcommand{\etal}{\textit{et al}.}
\newcommand{\ie}{\textit{i}.\textit{e}.}
\newcommand{\eg}{\textit{e}.\textit{g}.}
\newcommand{\mdegr}{360{\degree} }
\newcommand{\mk}[1]{{#1}}

\graphicspath{{fig/}}
\begin{document}

\title{Scanpath Prediction in Panoramic Videos via Expected Code Length Minimization}

\author{Mu~Li,
        Kanglong~Fan,
        and~Kede~Ma,~\IEEEmembership{Senior Member,~IEEE}% <-this % stops a space
\IEEEcompsocitemizethanks{
\IEEEcompsocthanksitem{This project was supported in part by the National Natural Science Foundation of China under Grant No. 62472124, Shenzhen Colleges and Universities Stable Support Program under Grant No. GXWD20220811170130002, Major Project of Guangdong Basic and Applied Basic Research under Grant No. 2023B0303000010, and Hong
Kong ITC Innovation and Technology Fund (9440379 and
9440390).}
\IEEEcompsocthanksitem Mu Li is with the School of Computer Science and Technology, Harbin Institute of Technology, Shenzhen, China, 518055 (e-mail: limuhit@gmail.com).
\IEEEcompsocthanksitem Kanglong Fan and Kede Ma (corresponding author) are with the Department
of Computer Science, City University of Hong Kong, Kowloon, Hong Kong (e-mail: kanglofan2-c@my.cityu.edu.hk, kede.ma@cityu.edu.hk).}% <-this % stops a space
% \thanks{Manuscript received xxx; revised xxx}
}

% The paper headers
\markboth{Submitted to IEEE Transactions on Pattern Analysis and Machine Intelligence}%
{Shell \MakeLowercase{\textit{et al.}}: A Sample Article Using IEEEtran.cls for IEEE Journals}

%\IEEEpubid{0000--0000/00\$00.00~\copyright~2021 IEEE}

\maketitle
\begin{abstract}
Scanpath prediction in panoramic videos is a challenging task due to the spherical geometry and multimodality of the input, and the inherent uncertainty and diversity of the output. To give a complete treatment of these characteristics, we first present a simple criterion for scanpath prediction based on principles from lossy data compression. This criterion suggests minimizing the expected code length of \textit{quantized} scanpaths, corresponding to fitting a \textit{discrete} conditional probability model via maximum likelihood. We condition the probability model on two modalities: a viewport sequence as the deformation-reduced visual input and a set of \textit{relative} past scanpaths projected onto respective viewports as the aligned path input. Furthermore, we parameterize it by a product of \textit{discretized} Gaussian mixture models to capture the uncertainty and diversity of scanpaths from different humans. In doing so, the training of the probability model does not rely on the specification of ``ground-truth'' scanpaths for imitation learning. We also introduce a proportional–integral–derivative (PID) controller-based sampler to generate realistic human-like scanpaths from the learned probability model. Experimental results demonstrate that our method consistently produces better quantitative scanpath results in terms of prediction accuracy (by comparing to the assumed ``ground-truths'') and perceptual realism (through machine discrimination) over a wide range of prediction horizons. We additionally verify the perceptual realism improvement via a formal psychophysical experiment and the generalization improvement on several unseen panoramic video datasets.
\end{abstract}


\section{Introduction}\label{sec:intro}

\IEEEPARstart{P}{anoramic} videos (also known as omnidirectional, spherical, and \mdegr videos) are gaining increasing popularity owing to their ability to provide more immersive viewing experience.
However, streaming and rendering \mdegr videos with minimal delay for real-time immersive and interactive experience remains a challenge due to the big data volume involved. To address this, viewport-adaptive streaming solutions have been developed, which transmit portions of the video in users' field of views (FoVs) at the highest possible quality while streaming the rest at lower quality to save bandwidth. These solutions depend crucially on accurate prediction of users' future viewpoints, typically in the form of visual saliency maps~\cite{zhu2019prediction,zhu2021viewing} or scanpaths~\cite{noton1971scanpaths,noton1971scanpathsineye}. A saliency map is an image that highlights the regions that are most likely to attract human attention, computed as the spatial aggregation of human scanpaths at a particular time instance. In contrast, a scanpath is represented as a time series of head/eye movement coordinates. Here, we focus on scanpath prediction, which provides an effective means of studying and summarizing viewing behaviors of a group of users when watching \mdegr videos with a broad range of applications, including panoramic video production~\cite{perazzi2015panoramic,zoric2013panoramic}, compression~\cite{ng2005data,cai2022overview}, processing~\cite{xu2017subjective,sitzmann2018saliency}, and rendering~\cite{rhee2017mr360,lee2017high}.

In the past decade, many scanpath prediction methods in \mdegr videos have been proposed, differing mainly in three aspects: 1) the input format and modality, 2) computational prediction mechanism, and 3) loss function. For the \textit{input format and modality}, Rond\'{o}n~\etal~\cite{rondon2022track} revealed that the user's past scanpath solely suffices to inform prediction for time horizons shorter than two to three seconds. 
Nevertheless, the majority of existing methods take \mdegr video frames as an ``indispensable'' form of visual input. Among numerous \mdegr video representations, the equirectangular projection (ERP) format is the most widely adopted, which, however, exhibits noticeable geometric deformations, especially for objects at high latitudes. For the \textit{computational prediction mechanism}, existing methods are inclined to rely on external algorithms for saliency prediction~\cite{fan2017fixation,xu2018gaze,rondon2022track} or optical flow estimation~\cite{fan2017fixation,xu2018gaze}, making the performance ``upper-bounded'' by these external methods, often trained on planar rather than \mdegr videos. After multimodal feature extraction and aggregation, a sequence-to-sequence predictor,  implemented by an unfolded recurrent neural network or a Transformer, is adopted to gather historical information. For the \textit{loss function}, some form of ``ground-truth'' scanpaths is commonly specified to gauge the prediction accuracy. A convenient choice is the mean squared error (MSE)~\cite{nguyen2018your,xu2018gaze,rondon2022track} or its spherical derivative~\cite{xu2022spherical}, which assumes the underlying probability distribution to be unimodal Gaussian. Such imitation learning is weak at capturing the scanpath uncertainty of an individual user and the scanpath diversity of different users. The binary cross-entropy (BCE)~\cite{fan2017fixation,li2018twolayer} between the predicted probability map of the next viewpoint and the normalized saliency map (aggregated from multiple ground-truth scanpaths) alleviates the diversity issue in the short term, but may lead to unnatural and inconsistent long-term predictions. In addition, auxiliary tasks such as fixation duration prediction~\cite{sun2021visual} and adversarial training~\cite{assens2018pathgan,martin2022scangan360} may be incorporated, further complicating the overall loss calculation and optimization.

\begin{figure}[!tbp]
\begin{minipage}{1\linewidth}
\begin{minipage}{0.48\linewidth}
\begin{tikzpicture}
\node (nd1) at (0,1.3) {\includegraphics[width=0.95\linewidth]{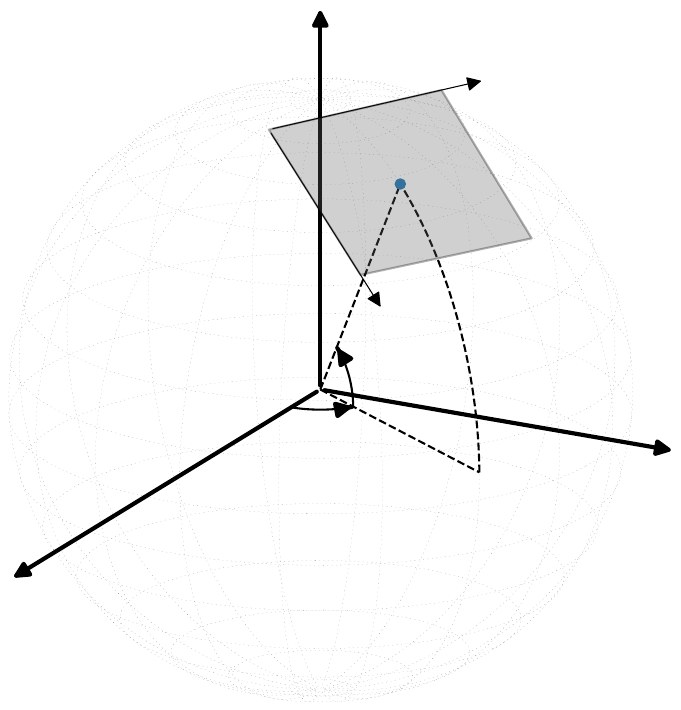}};
\node[text width=0.5cm,align=center] (t1) at (0.3,1.2) {\baselineskip=3pt \scriptsize{$\phi$} \par};
\node[text width=0.5cm,align=center] (t2) at (0,0.8) {\baselineskip=3pt \scriptsize{$\theta$} \par};
\node[text width=0.5cm,align=center] (t3) at (2,1) {\baselineskip=3pt \scriptsize{$y$} \par};
\node[text width=0.5cm,align=center] (t3) at (0.2,3.3) {\baselineskip=3pt \scriptsize{$z$} \par};
\node[text width=0.5cm,align=center] (t3) at (-1.8,-0.2) {\baselineskip=3pt \scriptsize{$x$} \par};
\node[text width=0.5cm,align=center] (t3) at (0.6,3.0) {\baselineskip=3pt \scriptsize{$u$} \par};
\node[text width=0.5cm,align=center] (t2) at (0,1.8) {\baselineskip=3pt \scriptsize{$v$} \par};
\end{tikzpicture}
\end{minipage}
\begin{minipage}{0.48\linewidth}
\begin{tikzpicture}
\node (nd1) at (0,1.3) {\includegraphics[width=0.8\linewidth]{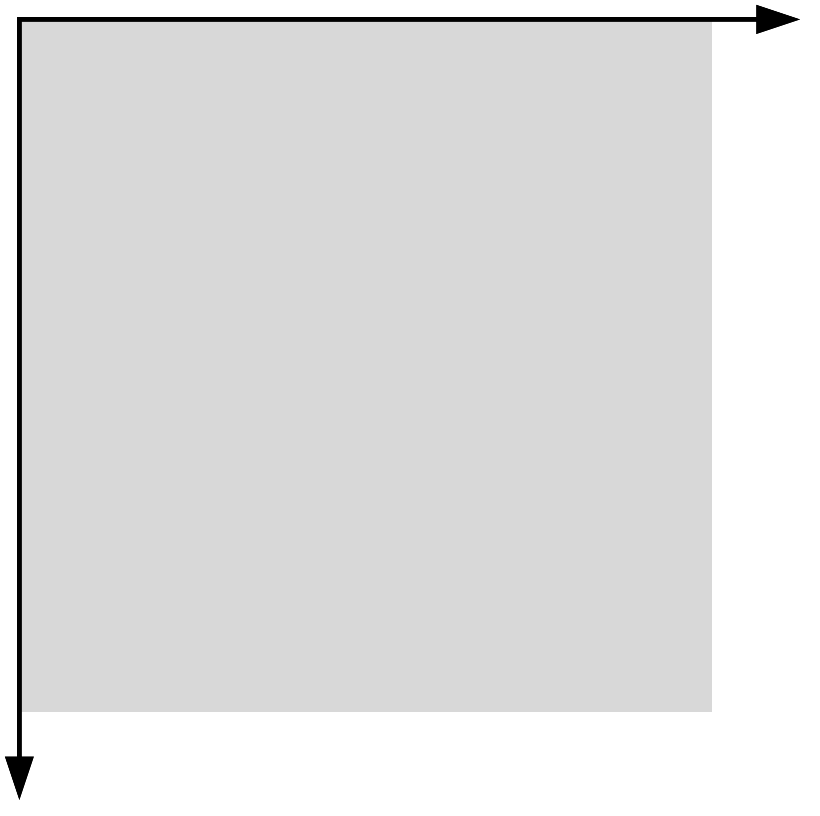}};
\node[text width=0.5cm,align=center] (t3) at (1.5,3.2) {\baselineskip=3pt \scriptsize{$u$} \par};
\node[text width=0.5cm,align=center] (t3) at (-1.8,-0.2) {\baselineskip=3pt \scriptsize{$v$} \par};
\end{tikzpicture}
\end{minipage}
\end{minipage}

\vspace*{0.3cm}

\begin{minipage}{1\linewidth}
\begin{minipage}{0.48\linewidth}
\centering{(a)}
\end{minipage}
\begin{minipage}{0.48\linewidth}
\centering{(b)}
\end{minipage}
\end{minipage}
\caption{Comparison of different coordinate systems used in \mdegr video processing. (a) Spherical coordinates $(\phi, \theta)$ and 3D Eculidean coordinates $(x, y, z)$. (b) Relative $uv$ coordinates $(u, v)$. }\label{fig:sphere2vp}
\end{figure}

In this paper, we address the problem of scanpath prediction through the lens of lossy data compression~\cite{cover2012elements}, where a scanpath is initially transformed and quantized into a discrete code, and subsequently compressed into a bitstream using entropy coding. 
% The bitstream length (\ie, the code length) can be estimated by the negative log probability of the discrete code. 
This leads to a simple new criterion---minimizing the expected code length---to learn an effective \textit{discrete} conditional probability model for scanpaths, from which we are able to sample realistic human-like scanpaths. Specifically, we condition our probability model on two modalities: the past \mdegr video frames and associated scanpath. To conform to the spherical nature of \mdegr videos, we choose to sample along the scanpath a sequence of rectilinear projections of viewports as the geometric deformation-reduced visual input compared to the ERP format. We further align the visual and positional modalities by projecting the scanpath (represented by spherical coordinates) onto each of the viewports (represented by \textit{relative} $uv$ coordinates, see Fig.~\ref{fig:sphere2vp}). This allows us to better represent and combine the multimodal features~\cite{baltruvsaitis2018multimodal} and to make easier yet better scanpath prediction in relative $uv$ space.  To capture the uncertainty and diversity of scanpaths, we parameterize the conditional probability model by a product of discretized Gaussian mixture models (GMMs), whose weight, mean, and variance parameters are estimated using feed-forward deep neural networks (DNNs). As a result, the expected code length can be approximated by the empirical expectation of their negative log probabilities.

Given the learned conditional probability model of scanpaths, we need a computational procedure to draw samples from it. To reliably imitate human viewing behaviors using a head-mounted display (HMD), we propose a variant of ancestral sampling based on a proportional-integral-derivative (PID) controller~\cite{bellman2015adaptive}. Specifically, we assume a proxy viewer governed by Newton's laws of motion, whose viewing actions can be transparently modeled and interpreted. The viewer begins exploring from some initial viewpoint with some initial speed and acceleration. We then sample a position from the learned distribution as the next viewpoint, and feed it to the PID controller as the new target to adjust the acceleration. In this way, the proxy viewer is guided to view towards the sampled viewpoint. By repeatedly sampling future viewpoints and adjusting the acceleration, we are able to generate human-like scanpaths of arbitrary length.

In summary, we make four primary contributions.

\begin{itemize}
\item We identify a neat criterion for scanpath prediction---expected code length minimization---which establishes the conceptual equivalence between scanpath prediction and lossy data compression.

\item We represent both visual and path contexts using relative $uv$ coordinates, thereby transforming panoramic scanpath prediction into a planar problem that is more tractable for computational modeling.

\item We develop a PID controller-based sampler to draw realistic, diverse, and long-horizon scanpaths, showing clear advantages over existing samplers. 

\item We conduct extensive experiments to quantitatively demonstrate the superiority of our method in terms of prediction accuracy (by comparing to ``ground-truths'') and perceptual realism (through machine discrimination and psychophysical testing) over different prediction horizons. We additionally verify the generalization of our method on several unseen panoramic video datasets.

\end{itemize}

\section{Related Work}

In this section, we review current scanpath prediction methods
in planar images, 360° images, and 360° videos, respectively,
thereby contextualizing our contributions.

\subsection{Scanpath Prediction in Planar Images}

Scanpath prediction was first investigated in planar images as a generalization of non-ordered prediction of eye fixations in the form of a 2D saliency map.  Ngo and Manjunath~\cite{ngo2017saccade} used a long short-term memory to process features extracted from a DNN for saccade sequence prediction. Wloka~\etal~\cite{wloka2018active} extracted and combined saliency information in a biologically plausible way for next fixation prediction together with a historical map of previous fixations. In contrast, Xia~\etal~\cite{xia2019predicting} constrained the DNN input to be localized to the current predicted fixation solely. Sun~\etal~\cite{sun2021visual} explicitly modeled the inhibition of return\footnote{Inhibition of return is defined as the relative suppression of processing of (detection of, orienting toward, or responding to) stimuli (object and events) that had recently been the focus of attention~\cite{klein1988inhibitory}.} mechanism when predicting the fixation location and duration. GMM was adopted for probabilistic modeling of the next fixation. A similar work~\cite{debelen2022scanpathnet} investigated the same mechanism, inspired by the Guided Search 6~\cite{wolfe2021guided}, a theoretical model of visual search in cognitive neuroscience.

\subsection{Scanpath Prediction in \mdegr Images and Videos}

\begin{table*}
\scriptsize
 \centering 
 \caption{Comparison of scanpath predictors in terms of the input format and modality, computational prediction mechanism and capability, and loss function.
NLL: negative log likelihood. BCE: binary cross-entropy loss. DTW: dynamic time warping loss. ``---'' means not available or applicable} \label{table1} 
 \resizebox{\linewidth}{!}{%
 \begin{tabular}{l l c l c c c}
     \toprule
     Method & Input Format \& Modality & External Algorithm & Sampling Method &  Horizon & GT & Loss\\
     \midrule
     Ngo17~\cite{ngo2017saccade} & planar image & ---  & beam search & --- & No & NLL\\ 
     Wloka18~\cite{wloka2018active} & planar image, past scanpath & saliency~\cite{Bruce2009Saliency, Huang2015SALICON}  & maximum likelihood & --- & No & --- \\   
     Xia19~\cite{xia2019predicting} & planar image, past scanpath & ---  & maximum likelihood & --- & Yes & BCE \\
     Sun21~\cite{sun2021visual} & planar image, past scanpath & instance segmentation~\cite{He2017Mask} & beam search & --- & No & NLL \\
     Belen22~\cite{debelen2022scanpathnet} & planar image, past scanpath & --- &  random sampling & --- & Yes & BCE \\
     Assens17~\cite{assens2017saltinet} & \mdegr image in ERP & --- & maximum likelihood & --- & Yes & BCE \\
     Zhu18~\cite{zhu2018prediction} & \mdegr image in viewport \& ERP & object detection~\cite{Felzenszwalb2010Object} & clustering \& graph cut & --- & No & --- \\
     Assens18~\cite{assens2018pathgan} & planar /\mdegr image in ERP & --- & random sampling & --- & Yes & MSE \& GAN \\
     Martin22~\cite{martin2022scangan360} & \mdegr image in ERP & --- & feed-forward generation & --- & Yes & DTW \& GAN \\
     Kerkouri22~\cite{kerkouri2022salypath360} & \mdegr image in ERP & saliency~\cite{dahou2020atsal} & maximum likelihood & --- & Yes & MSE \\[0.05cm]
     \midrule
     Fan17~\cite{fan2017fixation} & \mdegr video in ERP, past scanpath & saliency~\cite{cornia2016deep}, optical
     flow~\cite{lucas1981iterative} & probability thresholding & $1$ s & Yes & BCE\\ 
     Li18~\cite{li2018twolayer} & \mdegr video in ERP, past scanpath & saliency~\cite{deabreu2017look}, optical
     flow~\cite{lucas1981iterative} & probability thresholding & $1$ s & Yes & BCE\\
     Nguyen18~\cite{nguyen2018your} & \mdegr video in ERP, past scanpath & saliency~\cite{nguyen2018your} & maximum likelihood & $2.5$ s & Yes & MSE\\
     Xu18~\cite{xu2018gaze} & \mdegr video in ERP, past scanpath & saliency~\cite{pan2016shallow}, optical flow~\cite{ilg2017flownet} & maximum likelihood & $1$ s & Yes & MSE\\
     Xu19~\cite{xu2019predicting} & \mdegr video in viewport, past scanpath & --- & maximum  likelihood & $30$ ms & Yes & MSE\\
     Li19~\cite{li2019very} & past \& future scanpaths (from others) & saliency~\cite{Cornia2018predicting} (optional) & maximum likelihood & $10$ s & Yes & MSE\\
     TRACK~\cite{rondon2022track} & \mdegr video in ERP, past scanpath & saliency~\cite{nguyen2018your} & maximum likelihood & $5$ s & Yes & MSE\\
     VPT360~\cite{chao2021transformer} & past scanpath & --- & maximum likelihood & $5$ s & Yes & MSE\\
     Xu22~\cite{xu2022spherical} & \mdegr video in ERP, past scanpath & saliency~\cite{pan2016shallow}, optical flow~\cite{ilg2017flownet} & maximum likelihood & $1$ s & Yes & spherical MSE\\
     \midrule
     Ours & \mdegr video in viewport, past scanpath & --- & PID controller-based  & $\ge 20$ s & No & expected code length\\
     \bottomrule
    \end{tabular}%
    }
 \end{table*}

For scanpath prediction in \mdegr images, Assens~\etal~\cite{assens2017saltinet} proposed the concept of ``saliency volume'' as a sequence of time-indexed saliency maps in the ERP format. Scanpaths can be sampled from the predicted saliency volume based on maximum likelihood with inhibition. Zhu~\etal~\cite{zhu2018prediction} clustered and organized the most salient areas into a graph. Scanpaths were generated by maximizing the graph weights. 
Later, they~\cite{zhu2020learning} introduced deep reinforcement learning to predict head movements (HM) based on viewport images.
Assens~\etal~\cite{assens2018pathgan} combined MSE with an adversarial loss to encourage realistic scanpath generation. Similarly, Martin~\etal~\cite{martin2022scangan360} trained a generative adversarial network (GAN) with MSE replaced by a dynamic time warping-based loss~\cite{muller2007dynamic}. Kerkouri~\etal~\cite{kerkouri2022salypath360} adopted a differentiable  \texttt{argmax} surrogate to sample fixations memorylessly, with saliency prediction as an auxiliary task. 
\par
For scanpath prediction in \mdegr videos, Fan~\etal~\cite{fan2017fixation} combined saliency maps, optical flow maps, and historical viewing data (in the form of scanpaths or tiles\footnote{An ERP image can be divided into a set of nonoverlapping rectangular patches as tiles. Any FoV can be covered by a subset of tiles.}) to calculate the tile probabilities in future frames. Built upon~\cite{fan2017fixation}, Li~\etal~\cite{li2018twolayer} added a correction module to check and correct outlier tiles. Nguyen~\etal~\cite{nguyen2018your} improved panoramic saliency detection with the creation of a new \mdegr video saliency dataset. Similarly, Xu~\etal~\cite{xu2018gaze} improved saliency detection from a multi-scale perspective, and adopted \textit{relative} viewport displacement prediction but applied Euclidean geometry to spherical coordinates.
Xu~\etal~\cite{xu2019predicting} used deep reinforcement learning to imitate human scanpaths, limiting the prediction horizon to $30$ ms (\ie, one frame). Li~\etal~\cite{li2019very} leveraged not only the historical scanpath of the current user but also the full scanpaths of other users who had previously explored the same \mdegr video (as a form of cross-user behavior analysis). Importantly, Rond\'{o}n~\etal~\cite{rondon2022track} performed a thorough root-cause analysis of existing scanpath predictors. They identified that visual features only start contributing for horizons longer than two to three seconds, and recurrent visual feature analysis is crucial before concatenating with positional features. To respect the spherical nature of \mdegr videos, spherical convolution~\cite{cohen2018spherical,esteves2018learning,jiang2019spherical} has been adopted~\cite{wu2020spherical,xu2022spherical} in combination with spherical MSE as the loss function.
Additionally, Chao~\etal~\cite{chao2021transformer} explored a Transformer~\cite{vaswani2017attention,devlin2019bert} for scanpath prediction using its history as the sole input.
%\par

Table~\ref{table1} contrasts our scanpath predictor with existing representative ones in terms of the input format and modality, reliance on external algorithms, sampling method for the next viewpoint, prediction horizon, specification of ground-truth scanpaths, and loss function. It is clear that most existing panoramic scanpath predictors work directly with the ERP format for computational simplicity. Like~\cite{xu2019predicting}, we choose to sample a sequence of 2D viewports as the visual input, and further project the scanpath onto each of the viewports for relative scanpath prediction, both mitigating ERP-induced geometric deformations. Moreover, nearly all panoramic scanpath predictors take a \textit{supervised learning} approach: first specify ground-truth scanpaths, and then adopt MSE to quantify prediction errors, which is essentially equivalent to sampling the next viewpoint by maximizing unimodal Gaussian likelihood. Such a supervised learning formulation is limited to capturing the uncertainty and diversity of scanpaths. Interestingly, early work on planar scanpath prediction suggests taking an \textit{unsupervised learning} approach: first specify a parametric probability model of scanpaths, and then estimate its parameters through negative log likelihood (NLL) minimization. In a similar spirit, we optimize a probability model of panoramic visual scanpaths, as specified by a product of \textit{discretized} GMMs, by minimizing their expected code length. Our sampling strategy is also different and physics-driven. Additionally, our method can be end-to-end optimized, and does not rely on any external algorithm for visual feature analysis. 
\par
\section{Panoramic Scanpath Prediction}
\begin{figure*}[!tbp]
\scriptsize
\centering
\begin{tikzpicture}
\node (nd1) at (0,1.3) {\includegraphics[width=1\linewidth]{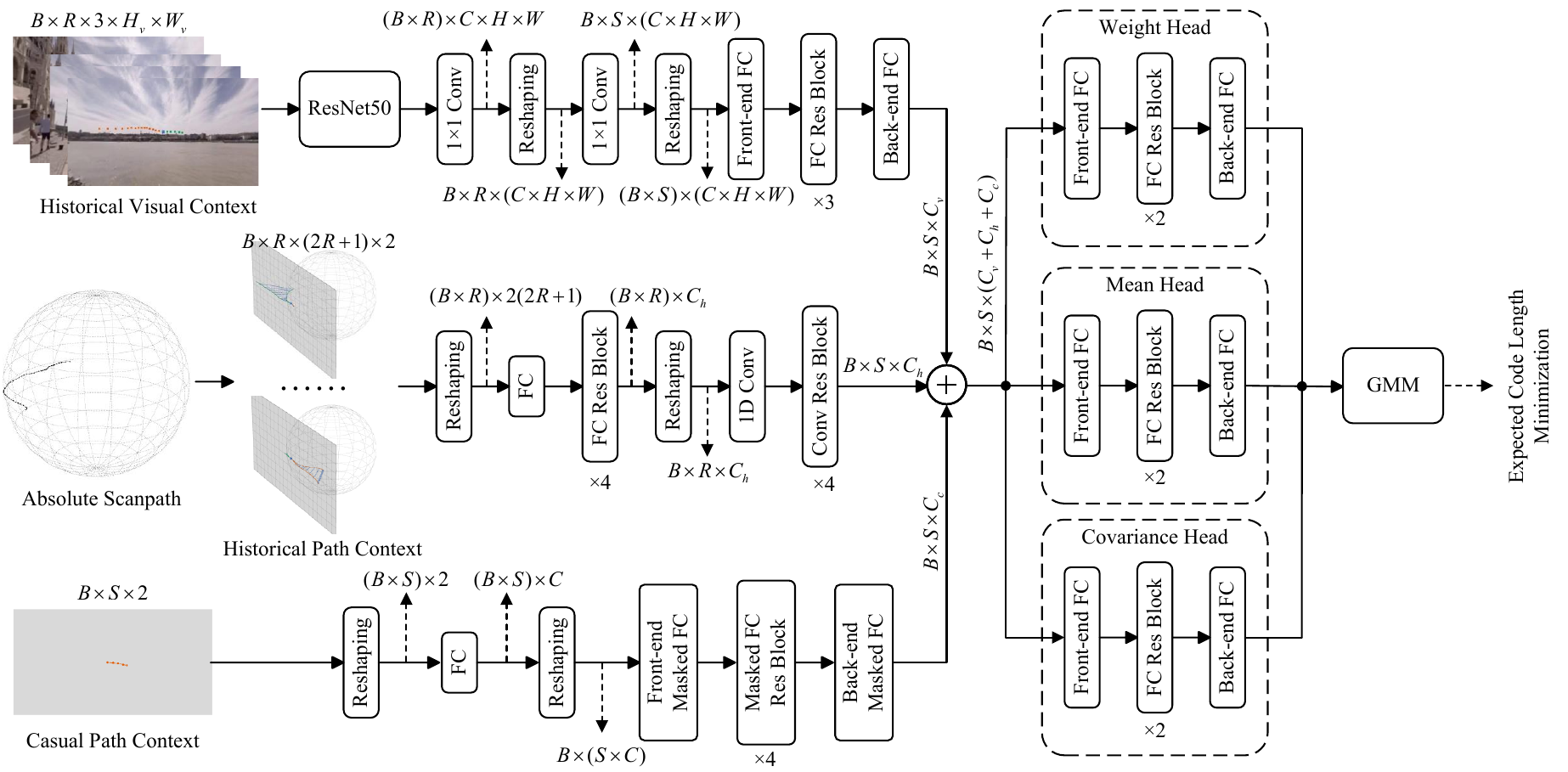}};
\node[draw,text width=2.2cm,align=center] (t1) at (7.8,-2.6) {\baselineskip=3pt {$\bigoplus$: Concatenation} \par};
\end{tikzpicture}
\vspace{-0.6cm}
\caption{System diagram of our discretized probability model for panoramic scanpath prediction. Initially, we extract features from the historical visual context, historical path context, and causal path context using DNNs. We then concatenate and feed these features to three prediction heads,  generating GMM parameters for future viewpoint prediction. $B$, $R$, and  $S$ represent the minibatch size, historical viewport number, and prediction horizon, respectively.}
\label{fig:SPNet}
\end{figure*}

In this section, we first formulate scanpath prediction from a probabilistic perspective, and connect it to lossy data compression. We then build our probability model on the historical visual and path contexts in the relative $uv$ space. Finally, we introduce the expected code length as the optimization objective for scanpath prediction. We summarize the proposed probabilistic scanpath predictor in Fig.~\ref{fig:SPNet}.

\subsection{Problem Formulation}
\label{subsec:pf}
Panoramic scanpath prediction aims to learn a sequence-to-sequence mapping $f:\{\mathcal{X},\bm{s}\} \mapsto {\bm r}$, in which a sequence of previous \mdegr video frames $\mathcal{X} = \{ \bm{x}_0, \ldots, \bm{x}_t,\ldots, \bm{x}_{T-1}\}$ and a sequence of historical viewpoints (\ie, the past scanpath) $\bm{s} = \{(\phi_0,\theta_0), \ldots, (\phi_t,\theta_t),\ldots,(\phi_{T-1},\theta_{T-1})\}$ are used to predict a sequence of future viewpoints (\ie, the future scanpath) $\bm{r}= \{ (\phi_T,\theta_T), \ldots, (\phi_{T+S-1},\theta_{T+S-1})\}$. Here, $S$ is the discrete prediction horizon; $(\phi_t,\theta_t)$ specifies the $t$-th viewpoint in the format of (latitude, longitude), which can be transformed into other coordinate systems (see Fig~\ref{fig:sphere2vp}); $\bm{x}_t$ denotes the $t$-th \mdegr video frame in any format, and here we obtain its viewport by first inverting the plane-to-sphere mapping followed by rectilinear projection centered at $(\phi_t,\theta_t)$.

A supervised learning approach to panoramic scanpath prediction relies on the specification of the ground-truth scanpath $\bm r$ to guide the optimization of the predictor $f$:
\begin{align}\label{eq:spf}
    \min D\left(f(\mathcal{X}, \bm s\right), \bm r),
\end{align}
where $D(\cdot, \cdot)$ is a distance measure between the predicted and ground-truth scanpaths. It is clear that Problem~\eqref{eq:spf} induces deterministic prediction, which falls short of modeling the scanpath uncertainty and diversity.

Inspired by early work on planar scanpath prediction~\cite{ngo2017saccade,sun2021visual} and optical flow estimation~\cite{simoncelli1993distributed}, we argue that it is preferred to formulate panoramic scanpath prediction as an unsupervised density estimation problem: 
\begin{align}
 \max p(f(\mathcal{X}, {\bm s}))=\max p(\bm r\vert\mathcal{X}, \bm s).
\end{align} 
We further decompose $p(\bm{r}\vert\mathcal{X}, \bm s)$ into the product of conditional probabilities of each viewpoint using the chain rule:
\begin{align}\label{eq:crop}
&&p(\bm{r}\vert \mathcal{X}, \bm{s}) =\prod_{t=0}^{S-1} p\left({\phi}_{T+t},{\theta}_{T+t} \Big\vert \mathcal{X}, \bm{s},\bm c_{t} \right),
\end{align}
where $\bm c_{t} = \{({\phi}_{T},{\theta}_{T}), ({\phi}_{T+1},{\theta}_{T+1}), \ldots, ({\phi}_{T+t-1},{\theta}_{T+t-1})\}$ and $\bm c_{0} = \emptyset$.  
The set of $\{\mathcal{X},\bm{s},\bm{c}_{t}\}$ constitutes the contexts of $({\phi}_{T+t},{\theta}_{T+t})$, among which $\mathcal{X}$ is the historical visual context, $\bm s$ is the historical path context, and  $\bm c_{t}$ is the causal path context, respectively. We may as well keep track of only the most recent visual and path contexts by placing a context window of size $R$. For the causal path context, we adopt human scanpaths during training, and sample them from the learned distribution during testing.

Generally,  estimating a probability density function in a high-dimensional space with a small finite sample set (as in our case) can easily lead to overfitting~\cite{bishop2006pattern}. We thus adopt quantization as a form of regularization with the goal of coarsing the estimated probability. From the \textit{computational} perspective, we introduce a hyperparameter---the quantization step size $\Delta$---that includes probability density estimation as a special case (\ie, $\Delta \rightarrow 0$). With a proper tuning of $\Delta$, a better (discretized) probability model for scanpath prediction shall be obtained (see ablation in Sec.~\ref{sec:ablation}). From the \textit{conceptual} perspective, optimizing a discretized probability model is deeply rooted in the well-established theory of lossy data compression, which offers a great opportunity to transfer recent advances in neural image compression~\cite{balle2016end,li2020efficient,li2021pseudocylindrical} to scanpath prediction.

A typical lossy data compression system consists of three major components: transformation, quantization, and entropy coding. As for panoramic scanpath prediction, the transformation step maps spherical coordinates in the form of $(\phi,\theta)$ into other coordinate systems such as the 3D Euclidean space~\cite{martin2022scangan360} and the relative $uv$ space adopted here. The quantization step truncates input values from a larger set (\eg, a continuous set) to output values in a smaller countable set with a finite number of elements. The uniform quantizer is the most widely used:
\begin{align}\label{eq:uq}
    Q(\xi) = \Delta\left\lfloor\frac{\xi}{\Delta} + \frac{1}{2}\right\rfloor,
\end{align}
where $\lfloor\cdot\rfloor$ denotes the floor function.

After quantization, we compute the discrete probability mass of $(\bar{\phi}_{T+t},\bar{\theta}_{T+t}) = (Q(\phi_{T+t}),Q(\theta_{T+t}))$ by accumulating the probability density defined in the righthand side of Eq.~\eqref{eq:crop} over the area $\Omega =[\bar{\phi}_{T+t}-1/2\Delta,\bar{\phi}_{T+t}+1/2\Delta]\times[\bar{\theta}_{T+t}-1/2\Delta,\bar{\theta}_{T+t}+1/2\Delta]$:
\begin{align}\label{eq:dprob}
P\left(\bar{\phi}_{T+t},\bar{\theta}_{T+t}\vert \mathcal{X}, \bm{s},\bm c_{t}\right) =  \int\limits_{\Omega} p\left(\bar{\phi}_{T+t},\bar{\theta}_{T+t} \vert \mathcal{X}, \bm{s},\bm c_{t} \right) \mathrm{d}\Omega.
\end{align}
Finally, given a minibatch of human scanpaths $\mathcal{B}=\{\mathcal{X}^{(i)},{\bm s}^{(i)}\}_{i = 1}^{B}$, where $\mathcal{X}^{(i)}=\{\bm x_0^{(i)},\ldots, \bm x_{T-1}^{(i)}\}$ and $\bm s^{(i)} = \{(\phi^{(i)}_0,\theta^{(i)}_0), \ldots, (\phi^{(i)}_{T-1}, \theta^{(i)}_{T-1})\}$, we may use stochastic optimizers~\cite{kingma2014adam} to minimize the NLL of the parameters:
\begin{equation}
\min -\frac{1}{B S}\sum_{i=1}^{B} \sum_{t=0}^{S-1} \log _2 \left(P\left(\bar{\phi}^{(i)}_{T+t},\bar{\theta}^{(i)}_{T+t}\Big\vert \mathcal{X}^{(i)}, \bm{s}^{(i)},\bm c^{(i)}_{t}\right)\right).
\end{equation}
It can be shown that this optimization is equivalent to minimizing the expected code length of training scanpaths, where $-\log_2\left( P\left(\bar{\phi}^{(i)}_{T+t},\bar{\theta}^{(i)}_{T+t}\Big\vert \mathcal{X}^{(i)}, \bm{s}^{(i)},\bm c^{(i)}_{t}\right)\right)$ provides a good approximation to the code length (\ie, the number of bits) used to encode $\left(\bar{\phi}^{(i)}_{T+t}, \bar{\theta}^{(i)}_{T+t}\right)$.

\subsection{Context Modeling}
\subsubsection{Historical Visual Context Modeling}\label{sec:visual_context}
Representing panoramic content in a plane is a long-standing problem that has been extensively studied in cartography. Unfortunately, there is no perfect sphere-to-plane projection, as stated in  Gauss's Theorem Egregium. Therefore, the question boils down to finding a panoramic representation that is less distorted and, meanwhile, more convenient to work with computationally. Instead of directly adopting the ERP sequence as the historical visual context, we resort to the viewport representation~\cite{xu2019predicting,sui2022perceptual}, which is less distorted and better reflects how users view \mdegr videos.

Specifically, a viewport $\bm x \in\mathbb{R}^{H_v\times W_v}$ with an FoV of $\phi_v \times \theta_v$ is defined as the tangent plane of a sphere, centered at the tangent point (\ie, the current viewpoint in the scanpath). To simplify the parameterization, we place the viewport (in $uv$ coordinates) on the plane $x=r$ centered at $[r,0,0]^\intercal$, where $r = 0.5 W_v \cot(0.5\theta_v)$ is the radius of the sphere. As a result, a pixel location $(u, v)$ in the viewport can be conveniently represented by $[r,y,z]^\intercal$ in the 3D Euclidean space, where $y = u - 0.5W_v + 0.5$ and $z=0.5H_v - v - 0.5$.
We rotate the center of the viewport to the current viewpoint $(\phi, \theta)$ using the Rodrigues' rotation formula, given an axis (described by a unit-length vector $\bm k \in \mathbb{R}^3 = [k_x, k_y, k_z]^\intercal$) and an angle of rotation $\omega$ (using the righthand rule):
\begin{align}\label{eq:rot1}
    {\bm q}^\mathrm{rot}  &= \mathrm{Rodrigues} (\bm q; \bm k, \omega)\nonumber\\
    &=(\mathbf I + \sin(\omega)\mathbf K + (1-\cos(\omega)\mathbf K^2) \bm q,
\end{align}
where $\bm q = [r,x,y]^\intercal$ and 
\begin{align}\label{eq:rot2}
   \mathbf{K} =  \begin{bmatrix}0 & -k_z & k_y\\
k_z & 0 & -k_x\\
-k_y&k_x&0
\end{bmatrix}.
\end{align}
We use Eq.~\eqref{eq:rot1} to first rotate a pixel location $\bm q = [r,y,z]^\intercal$ in the viewport with respect to the $z$-axis by $\theta$:
\begin{align}\label{eq:rot3}
      \bm q'= \mathrm{Rodrigues} (\bm q; [0,0,1]^\intercal, \theta),
\end{align}
and further rotate it with respect to the rotated $y$-axis 
\begin{align}\label{eq:rot4}
      \bm y' = \mathrm{Rodrigues} ([0, 1, 0]^\intercal; [0,0,1]^\intercal, \theta)
\end{align}
by $-\phi$:
\begin{align}\label{eq:rot5}
      \bm q^\mathrm{rot} = \mathrm{Rodrigues} (\bm q'; {\bm y}', -\phi).
\end{align}
The rotation process described in Eqs.~\eqref{eq:rot3} to~\eqref{eq:rot5} can be compactly expressed as
\begin{align}\label{eq:rotation}
\bm q^\mathrm{rot} = \mathbf{R}(\phi,\theta) \bm q,
\end{align}
where $\mathbf{R}(\phi,\theta)\in\mathbb{R}^{3\times 3}$ denotes the rotation matrix.
Finally, we transform $\bm q^\mathrm{rot}  =  [q^\mathrm{rot} _x,q^\mathrm{rot} _y,q^\mathrm{rot} _z]^\intercal$ back to the sphere:
\begin{align}\label{eq:theta_phi}
     \phi' = \arcsin(q^\mathrm{rot}_z/r) \quad \mbox{and} \quad \theta' = \arctan2(q^\mathrm{rot}_y / q^\mathrm{rot}_x),
\end{align}
where $\arctan2(\cdot)$ is the $2$-argument arctangent\footnote{\url{https://en.wikipedia.org/wiki/Atan2}}, and relate $(\phi',\theta')$ to the discrete sampling position $(m, n)$ in ERP:
\begin{align}\label{phitom}
    m = (0.5 - \phi'/\pi) H -0.5
\end{align}
and
\begin{align}\label{thetaton}
    n =(\theta'/2\pi +0.5)W-0.5.
\end{align}
With that, we complete the mapping from the $(u,v)$ coordinates in the viewport to $(m,n)$ coordinates in the ERP format.
 In case the computed $(m,n)$ according to Eqs.~\eqref{phitom} and~\eqref{thetaton} are non-integers, we interpolate their values bilinearly.  For each viewpoint, we generate a corresponding viewport, giving rise to the viewport sequence $\mathcal{X} = \{ \bm{x}_{T-R}, \ldots, \bm{x}_{T-1}\}$ that can be seen as a standard planar video clip.

 \begin{figure*}[!tbp]
\scriptsize
\centering
\begin{minipage}{0.24\textwidth}
\centering
\includegraphics[width=0.8\textwidth]{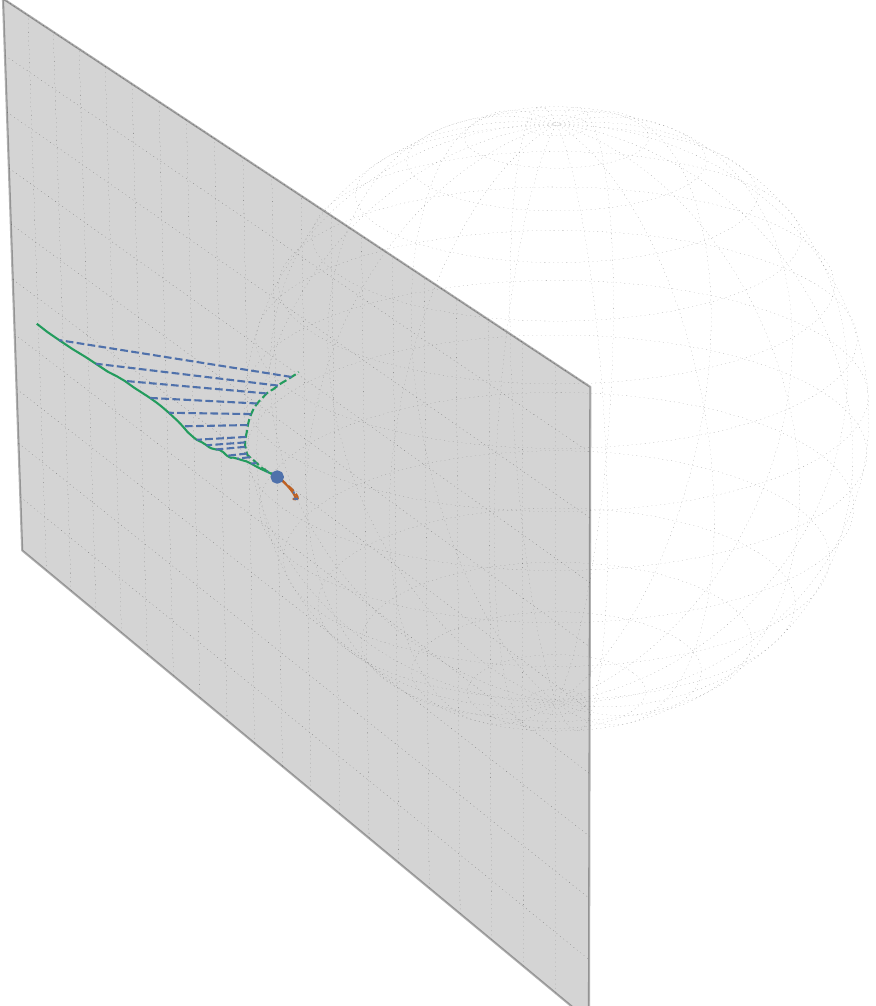}

\vspace{0.1cm}

\includegraphics[width=1\textwidth]{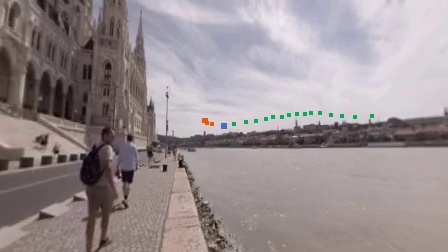}
\end{minipage}
\begin{minipage}{0.24\textwidth}
\centering
\includegraphics[width=0.8\textwidth]{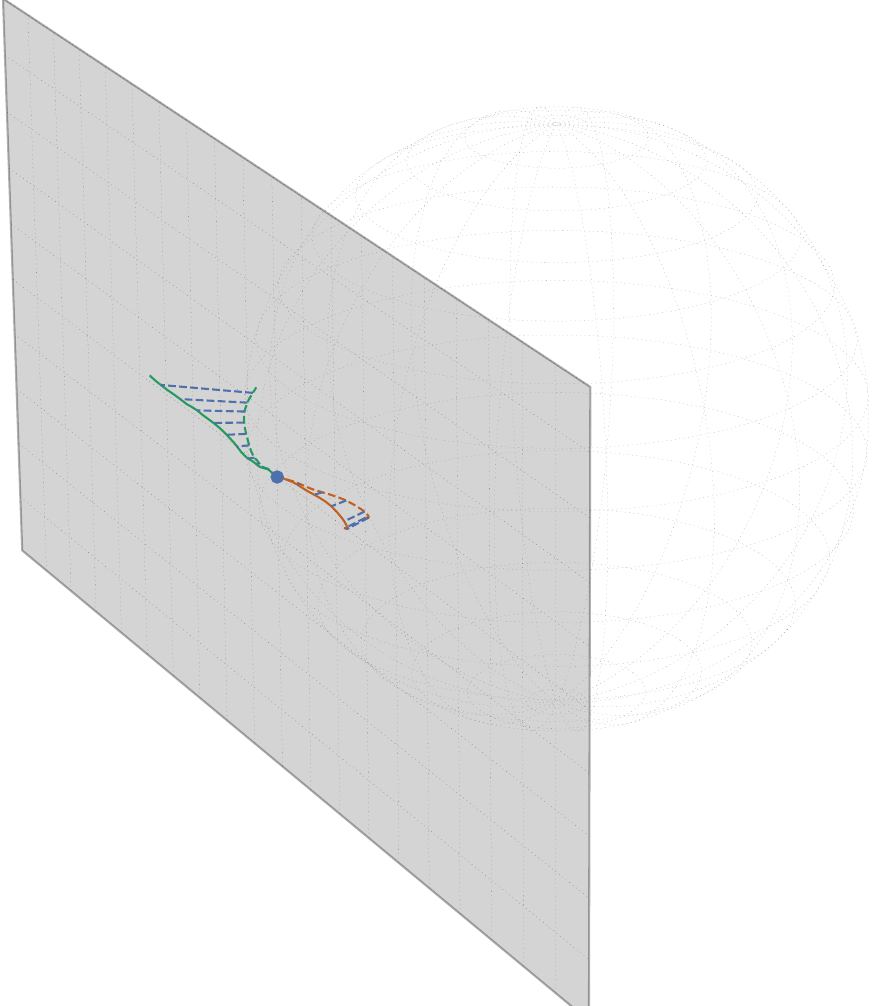}

\vspace{0.1cm}

\includegraphics[width=1\textwidth]{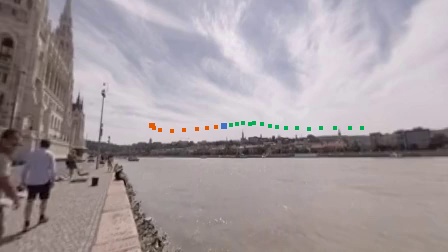}
\end{minipage}
\begin{minipage}{0.24\textwidth}
\centering
\includegraphics[width=0.8\textwidth]{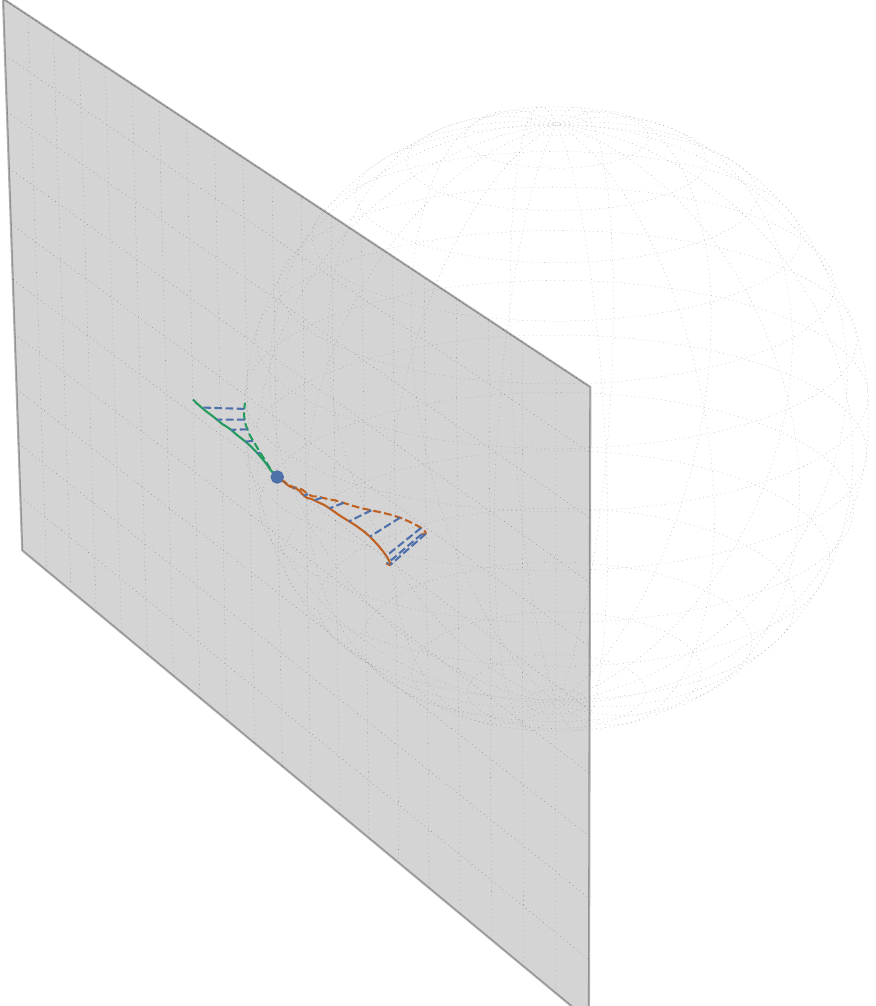}

\vspace{0.1cm}

\includegraphics[width=1\textwidth]{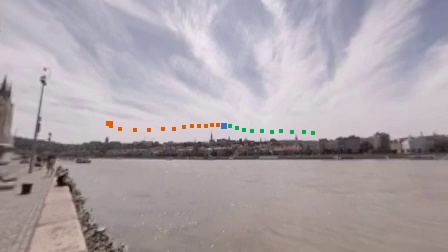}
\end{minipage}
\begin{minipage}{0.24\textwidth}
\centering
\includegraphics[width=0.8\textwidth]{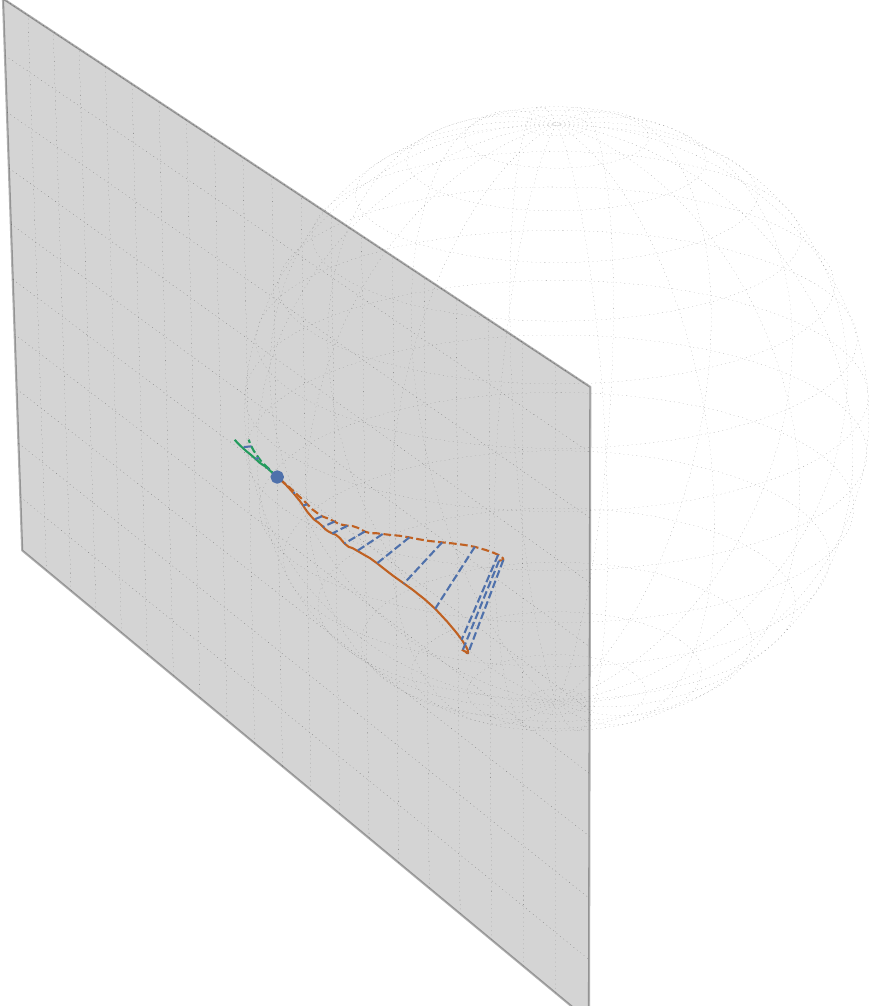}

\vspace{0.1cm}

\includegraphics[width=1\textwidth]{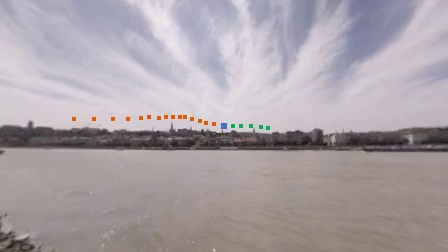}
\end{minipage}
\caption{Visualization of the projected scanpaths onto different viewports. The top row shows the procedure of projecting the same scanpath $\bm s$ onto different viewports. The orange (green) dots indicate the viewpoints before (after) the anchor blue viewpoint, from which we extract the anchor viewport for projection. The bottom row overlaps the projected scanpath and corresponding viewport, where we center the anchor viewpoint using Eq.~\eqref{eq:shifteduv}.}\label{fig:reluv}
\end{figure*}

To extract visual features for predicting $S$ future viewpoints from the set $ \{\mathcal{X}^{(i)}\}_{i=1}^{B} $, where each $\mathcal{X}^{(i)}$ includes $ R $ past viewports, we use a variant of ResNet50~\cite{he2016deep} by replacing the last global average pooling layer and the fully connected (FC) layer with a $1\times 1$ convolution layer for channel dimension adjustment. We stack the $R$ historical viewports in the batch dimension to parallelize spatial feature extraction, leading to an output representation of size $(B\times R )\times C\times H\times W$, where $C, H, W$ are the channel number,  height, and width, respectively. We then reshape the features to $B\times R \times (C\times H\times W)$, splitting the batch and time dimensions, and flattening spatial and channel dimensions. A 1D convolution is applied to adjust the time dimension to $S$ (\ie, the prediction horizon).   Finally, we reshape the features to $(B \times S) \times (C\times H\times W)$, and adopt a multilayer perceptron (consisting of one front-end FC layer, three FC residual blocks\footnote{The FC residual block is composed of two FC layers followed by layer normalization and leaky ReLU activation.}, and one back-end FC layer) to compute the final visual features of size $B\times S\times C_v$.

\subsubsection{Historical Path Context Modeling} Panoramic scanpaths have commonly been represented using spherical coordinates $(\phi, \theta)$ (or their discrete counterparts $(m,n)$) or 3D Euclidean coordinates $(x,y,z)$. However, these absolute coordinates are neither user-centric, meaning that the past and future viewpoints are not relative to the current viewpoint, nor well aligned with the visual context. To remedy both, we propose to represent the scanpath in the relative $uv$ space. As shown in Fig.~\ref{fig:reluv}, by projecting the historical scanpath $\bm s$ onto each of the viewports, we model the current viewpoint of interest and future viewpoints to be predicted from the viewer's perspective. Moreover, aligning data from different modalities has proven effective in multimodal tasks~\cite{baltruvsaitis2018multimodal}. Similarly, we align the visual and path contexts in the same $uv$ space, which benefits scanpath prediction and bridges the computational modeling gap between planar and panoramic videos.

 Given the anchor time stamp $t$, we extract the viewport tangent at $(\phi_t,\theta_t)$, and project the scanpath onto it, by inverse mapping from $(\phi, \theta)$ to $(u,v)$. Specifically, we first cast $(\phi_k, \theta_k)\in \bm s$ to 3D Euclidean coordinates:
 \begin{equation}
\begin{bmatrix}
x_k\\
y_k\\
z_k
\end{bmatrix}
= 
\begin{bmatrix}
r\cos(\phi_k)\cos(\theta_k) \\
r\cos(\phi_k)\sin(\theta_k) \\
r\sin(\phi_k)
\end{bmatrix},
\end{equation}
rotate $[x_k, y_k, z_k]^\intercal$ by the transpose of $\mathbf{R}(\phi_t,\theta_t)$:
\begin{equation}
\begin{bmatrix}
x_{tk}\\
y_{tk}\\
z_{tk}
\end{bmatrix}
= \mathbf{R}^\intercal(\phi_t, \theta_t) 
\begin{bmatrix}
x_k \\
y_k \\
z_k
\end{bmatrix},
\end{equation}
and project $[x_{tk},y_{tk},z_{tk}]^\intercal$ onto the plane $x=r$:
\begin{equation}
\begin{bmatrix}
x'_{tk}\\
y'_{tk}\\
z'_{tk}
\end{bmatrix}
= 
\begin{bmatrix}
r \\
y_{tk}\cdot r/ x_{tk} \\
z_{tk}\cdot r / x_{tk}
\end{bmatrix},
\end{equation}
where we add a subscript ``$t$'' to emphasize that the historical scanpath $\bm s$ is projected onto the $t$-th anchor viewport. We further convert $[x'_{tk},y'_{tk},z'_{tk}]^\intercal$ to $uv$ coordinates:
\begin{equation}
\begin{bmatrix}
u'_{tk}\\
v'_{tk}
\end{bmatrix}
= 
\begin{bmatrix}
 y'_{tk} + 0.5 W_v - 0.5 \\
0.5 H_v - z'_{tk} - 0.5
\end{bmatrix}.
\end{equation}
We last shift the $uv$ plane by moving the center of viewport from $(0.5W_v - 0.5, 0.5H_v-0.5)$ to $(0,0)$. The projection of $(\phi_k, \theta_k)$ onto the $t$-th viewport is then represented by
\begin{equation}\label{eq:shifteduv}
\begin{bmatrix}
u_{tk}\\
v_{tk}
\end{bmatrix}
= 
\begin{bmatrix}
 y'_{tk} \\
- z'_{tk}
\end{bmatrix},
\end{equation}
where $(u_{tt}, v_{tt}) = (0,0)$.

To extract historical path features for predicting $S$ future viewpoints from $\{\bm s^{(i)}\}_{i=1}^B$, we first reshape the input from $B\times R \times (2R+1) \times 2$, where $B,R,(2R+1)$ are, respectively, the minibatch size, the historical viewport number, and the context window size of the projected scanpaths, to $(B\times R)\times 2(2R+1)$, and process it with an FC layer and an FC residual block to obtain an intermediate output of size $(B\times R) \times C_h$. We then split the first two dimensions (\ie, $(B\times R)\times C_h \rightarrow B\times R\times C_h$), and append a 1D convolution layer and four 1D convolutional residual blocks\footnote{The 1D convolution residual block consists of two convolutions followed by batch normalization and leaky ReLU activation.} to produce the final historical path features of size $B \times S \times C_h$.

\subsubsection{Causal Path Context Modeling}
Similarly, we model the causal path context $\bm c_t$ by projecting it onto the anchor viewport $\bm x_{T-1}$, and use masked computation to ensure causal modeling. We first reshape the input from $B\times S\times 2$ to $(B\times S)\times 2$, and use an FC layer to transform the two-dimensional coordinates to a $C$-dimensional feature representation. We then stack the last two dimensions (\ie, $(B\times S)\times C \rightarrow B\times (S\times C)$), and apply a masked multilayer perceptron, consisting of a front-end masked FC layer, four masked FC residual blocks, and a back-end masked FC layer to compute the causal path features of size $B \times S \times C_c$. The masked FC layer is defined as
\begin{equation}
\bm{h}^\intercal = (\mathbf{M}\otimes \mathbf{W})  \bm{g}^\intercal,
\end{equation} 
where $\otimes$ is the Hadamard product, and $\bm{g} \in \mathbb{R}^{B \times  (S \times C_\mathrm{in})}$ and $\bm{h} \in \mathbb{R}^{B \times  (S \times C_\mathrm{out})}$ are the input and output features, respectively. $\mathbf{W}, \mathbf{M} \in \mathbb{R}^{(S\times C_\mathrm{out}) \times (S\times C_\mathrm{in})}$ are the weight and mask matrices, respectively, in which 
\begin{equation}
M_{ij} = \begin{cases}
1 & \text{if }\lfloor j/C_\mathrm{in} \rfloor < \lfloor i/C_\mathrm{out} \rfloor \\
0 &\text{otherwise},\\
\end{cases}
\end{equation}
for the front-end layer and
\begin{equation}
M_{ij} = \begin{cases}
1& \text{if } \lfloor j/C_\mathrm{in} \rfloor \leq \lfloor i/C_\mathrm{out}\rfloor \\
0& \text{otherwise},\\
\end{cases}
\end{equation}
for the hidden and back-end layers. 

\subsection{Objective Function}
Inspired by the entropy modeling  in the field of neural image compression~\cite{balle2016end,li2020efficient}, we construct the probability model of 
 $\bar {\bm \eta}_{T-1,t} = [\bar{u}_{T-1,t},\bar{v}_{T-1,t}]^\intercal$, the quantized version of ${\bm \eta}_{T-1,t} = [{u}_{T-1,t},{v}_{T-1,t}]^\intercal$, using a GMM with $K$ components. Our GMM is conditioned on  the historical visual context $\mathcal{X}$, historical path context $\bm s$, and causal path context $\bm c_t$:
\begin{align}\label{eq:qgmm}
 &\mathrm{GMM}\left(\bar{\bm \eta}_{t} \Big\vert \mathcal{X}, \bm{s}, \bm{c}_t; \bm \alpha, \{\bm \mu_k\}_{k=1}^K, \{\bm \Sigma_k\}_{k=1}^K\right) = \nonumber\\
 & \sum_{k=1}^K \frac{\alpha_k} {2\pi\sqrt{\vert\bm{\Sigma}_k\vert}} \exp\left( -\frac{1}{2} (\bar{\bm{\eta}}_t-\bm{\mu}_k)^\intercal \bm{\Sigma}_k^{-1}(\bar{\bm{\eta}}_t-\bm{\mu}_k)\right),
\end{align}
where we omit the subscript $T-1$ in $\bar{\bm \eta}_t$ to make the notations uncluttered. Due to the fact that the gradients of the quantizer in Eq.~\eqref{eq:uq} are zeros almost everywhere, we approximate it during training by adding a random noise $\epsilon$ uniformly sampled from $[-\Delta, \Delta]$ to the continuous value~\cite{balle2016end}:
\begin{align}
    Q(\xi;\epsilon) = \xi +\epsilon.
\end{align}
 $\{ \alpha_k, \bm \mu_k, \bm \Sigma_k\}$ in Eq.~\eqref{eq:qgmm} represent the estimated mixture weight, mean vector, and covariance matrix of the $k$-th Gaussian component, respectively. Such estimation can be made by concatenating the visual and path features (with the size of $B\times S\times (C_v+C_h+C_c)$), followed by three prediction heads. We assume the independence between the horizontal direction $u$ and vertical direction $v$, resulting in diagonal covariance matrices. Each prediction head consists of a front-end FC layer, two FC residual blocks, and a back-end FC layer. We append a softmax layer at the end of the weight prediction head to ensure a probability vector output. Similarly, we add ReLU at the end of the covariance prediction head to ensure nonnegative outputs on the diagonals.

We discretize the GMM model by integrating the probability density over the area $\Omega = [\bar{u}_{t}-1/2\Delta,\bar{u}_{t}+1/2\Delta]\times [\bar{v}_{t}-1/2\Delta,\bar{v}_{t}+1/2\Delta]$:
\begin{align}\label{eq:prob_acc}
P\left(\bar{\bm \eta}_t\Big\vert \mathcal{X}, \bm{s}, \bm{c}_t\right) =  \int\limits_{\Omega} \mbox{GMM}\left(\bar{\bm \eta}_t\Big\vert \mathcal{X}, \bm{s}, \bm{c}_t\right) \mathrm{d}\Omega.
\end{align}

Last, we end-to-end optimize the entire model by minimizing the expected code length of the scanpaths in a minibatch:
\begin{equation}\label{eq:obj}
\min -\frac{1}{B S}\sum_{i=1}^{B} \sum_{t=0}^{S-1} \log _2 \left(P\left(\bar{\bm \eta}_t^{(i)}\Big\vert \mathcal{X}^{(i)}, \bm{s}^{(i)},\bm c^{(i)}_{t}\right)\right).
\end{equation}

\section{PID Controller for Scanpath Sampling}\label{sec:sample}
Probabilistic scanpath prediction requires a sampler to draw future viewpoints from the learned probability model. Being causal (\ie, autoregressive), our probability model as a product of discretized GMMs fits naturally to ancestral sampling. That is, we start by initializing the causal path context to be an empty set and conditioning on the historical visual and path contexts to draw the first viewpoint. We put the sampled viewpoint into the causal path context for next viewpoint generation. By repeating this step, we are able to predict an $S$-length scanpath, which completes a \textit{sampling round}. We then update the historical visual context by extracting a sequence of $S$ viewports along the newly sampled scanpath, which is also used to override the historical path context. The causal path context is then cleared for the next round of scanpath prediction. By completing multiple rounds, our scanpath predictor supports very long-term (and in theory arbitrary-length) scanpath generation.

It remains to specify the sampler for next viewpoint generation. One straightforward instantiation is to draw a random sample from the distribution as the next viewpoint by inverse transform sampling~\cite{devroye2006nonuniform}. 
Empirically, this sampler tends to produce less smooth scanpaths, leading to shaky viewport sequences. Another option is to sample the next viewpoint that has the maximum probability mass. This sampler is closely related to directly regressing the next viewpoint in the supervised learning setting; as a result, it tends to generate similar, repeated scanpaths, and may even get stuck at a single position for an extended period.

 To address these issues, we propose to use a PID controller~\cite{ang2005pid} to guide the sampling procedure. The PID controller is a widely used feedback mechanism that allows for continuous modulation of control signals to achieve stable control. Here, we assume a proxy viewer based on Newton’s laws of motion. At the beginning, the proxy viewer is placed at the starting point $\hat{\bm\eta}_{-1}=[0,0]^\intercal$ in the $uv$ coordinate system, with the given initial speed $\bm b_{-1}$ and acceleration $\bm a_{-1}$. The $t$-th viewpoint is predicted by
 \begin{align}\label{eq:nm}
     \hat{\bm \eta}_t = \hat{\bm \eta}_{t-1} + \Delta \tau\bm b_{t-1} + \frac{1}{2}(\Delta\tau )^2\bm a_{t-1}, t\in \{0,\ldots, S-1\},
 \end{align}
where the speed $\bm b_{t-1}$ is updated by 
\begin{align}
    \bm b_t = \bm b_{t-1} + \Delta \tau\bm a_{t-1},
\end{align}
 and $\Delta \tau$ is the sampling interval (\ie, the inverse of the sampling rate). To update the accelartion $\bm a_{t-1}$, we first provide a reference viewpoint $\bar{\bm \eta}_t$ for $\hat{\bm \eta}_t$ by drawing a sample from $P\left(\bar{\bm \eta}_t\Big\vert \mathcal{X}, \bm{s}, \bm{c}_t\right)$, where $\bm c_t =\{\hat {\bm \eta}_0, \ldots, \hat{\bm \eta}_{t-1}\}$, using inverse transform sampling. An error signal can then be generated:
 \begin{align}
     \bm e_t = \bar{\bm \eta}_t - \hat{\bm \eta}_t,
 \end{align}
which is fed to the PID controller for acceleration adjustment:
 \begin{align}\label{eq:pida}
     \bm a_t = K_p \bm e_t + K_i \sum_{\tau=0}^{t}\bm e_\tau + K_d (\bm e_t - \bm e_{t-1}),
 \end{align}
where $K_p$, $K_i$, and $K_d$ are the proportional, integral, and derivative gains, respectively. One subtlety is that when we move to the next sampling round, we need to transfer and represent $\bm b_t$ and $\bm a_t$ in the new $uv$ space defined on the viewport at time stamp $T+S-1$ (rather than time stamp $T-1$). This can be done by keeping track of one more viewpoint using Eq.~\eqref{eq:nm} for speed and acceleration computation in the new $uv$ space. In practice, it suffices to transfer only the average speed (\ie, $(\hat{\bm \eta}_S - \hat{\bm \eta}_{S-1})/\Delta \tau$), and reset the acceleration to zero because the latter is usually quite small.

\section{Experiments}
In this section, we first describe the panoramic video datasets used as evaluation benchmarks, followed by the experimental setups. We then compare our method with existing panoramic scanpath predictors in terms of prediction accuracy, perceptual realism, and generalization on unseen datasets. We last conduct comprehensive ablation studies to justify the design choices of the proposed method. The trained models and accompanying code are available at  \url{https://github.com/limuhit/panoramic_video_scanpath}.

\begin{table}[!tbp]
 \centering 
 \caption{Summary of panoramic video datasets for scanpath prediction. In the last column,
 NP indicates natural photographic videos, while CG stands for computer-generated videos} \label{tab:dataset} 
\scriptsize
 \begin{tabular}{l| c c c p{1cm}}
     \toprule 
     Dataset & \#Videos & \#Scanpaths & Duration & Type \\
     \midrule 
     NOSSDAV17~\cite{fan2017fixation} & $10$ & $250$ & $60$ s & NP/CG \\
     ICBD16~\cite{bao2016shooting} & $16$ & $976$ & $30$ s & NP \\
     MMSys17~\cite{wu2017dataset} & $18$ & $864$ & $164$-$655$ s & NP \\
     MMSys18~\cite{david2018dataset} & $19$ & $1,083$ & $20$ s & NP \\
     PAMI19~\cite{xu2019predicting} & $76$ & $4,408$ & $10$-$80$ s & NP/CG \\
     CVPR18~\cite{xu2018gaze} & $208$ & $6,672$ & $20$-$60$ s & NP\\
     VRW23~\cite{fang2022subjective} & $502$ & $20,080$ & $15$ s & NP/CG \\
     \bottomrule
    \end{tabular}
 \end{table}
 
\subsection{Datasets}
Panoramic video datasets typically contain eye-tracking data, in the form of eye movements and head orientations, collected from human participants. We list some basic information of commonly used \mdegr video datasets in Table~\ref{tab:dataset}.
Based on the dataset scale, we have selected the CVPR18 dataset~\cite{xu2018gaze} and VRW23 dataset~\cite{fang2022subjective} for the main experiments, and leave some others for cross-dataset generalization testing. To illustrate the scanpath diversity in the two  datasets, we evaluate the consistency of two scanpaths of the same length using the temporal correlation:
\begin{equation}\label{TC}
    \mathrm{TC}\left(\bm s^{(i)},\bm s^{(j)}\right)=
    \frac{1}{2}\left(\mathrm{PCC}\left(\bm \phi^{(i)},\bm \phi^{(j)}\right) +\mathrm{PCC}\left(\bm \theta^{(i)},\bm \theta^{(j)}\right)\right),
\end{equation}
where $\mathrm{PCC}(\cdot)$ computes the Pearson correlation coefficient. The mean temporal correlation over $N$ scanpaths for the same \mdegr video can be computed by 
\begin{equation}\label{meanTC}
    \mathrm{meanTC}\left(\{\bm s^{(i)}\}_{i=1}^N\right) = \frac{\sum\limits_{i=1}^N\sum\limits_{j=i+1}^N\mathrm{TC}\left(\bm s^{(i)},\bm s^{(j)}\right)}{N(N-1)/2},
\end{equation}
which ranges from $[-1,1]$, with a larger value indicating higher temporal consistency.

\begin{figure}[!tbp]
\scriptsize
\centering
\includegraphics[width=0.40\textwidth]{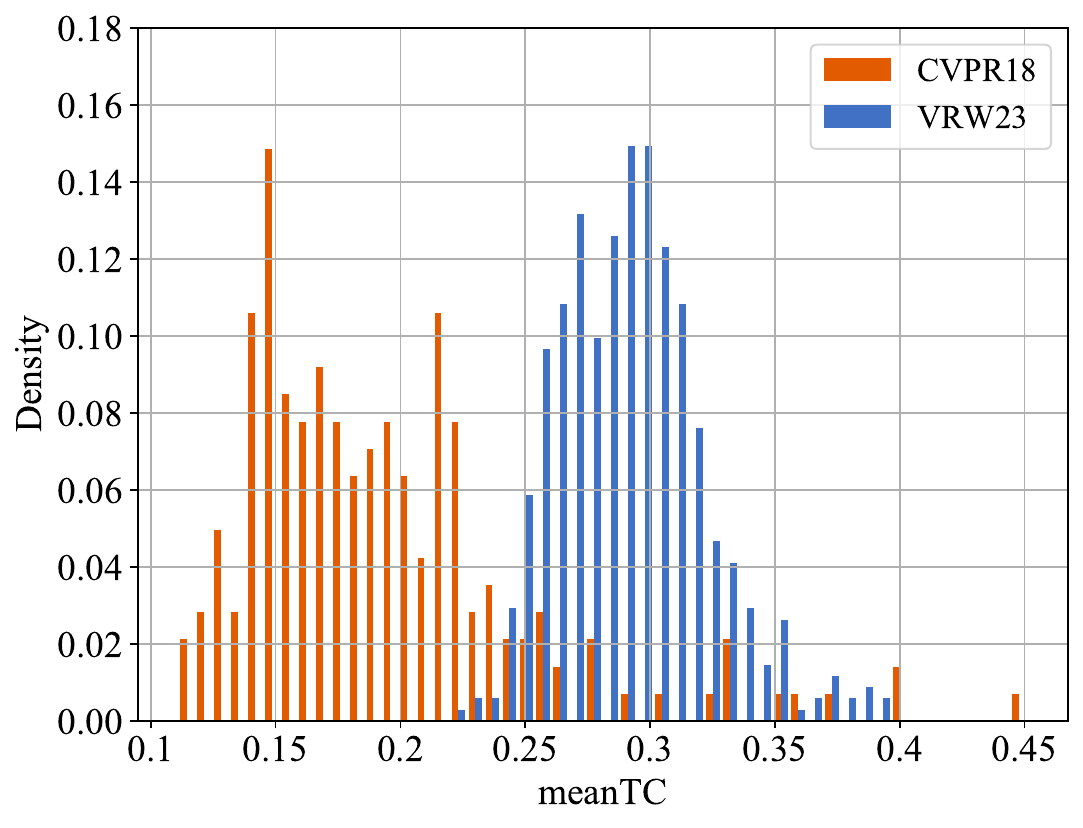}
\caption{meanTC histograms of the CVPR and VRW23 datasets.}
\label{fig:data_plcc}
\end{figure}

We visualize the meanTC histograms of the CVPR18 and VRW23 datasets in Fig.~\ref{fig:data_plcc}, from which we observe that scanpaths in both datasets exhibit considerable diversity, which shall be computationally modeled. Moreover, scanpaths with longer horizons in the CVPR18 dataset  (\eg, more than $30$ seconds) are even less consistent, showing the difficulty of long-term scanpath prediction.

\subsection{Experimental Setups}\label{subsec:expsetup}
In the main experiments, we follow~\cite{xu2018gaze,rondon2022track} to downsample both the video (and corresponding) scanpaths to five frames (and viewpoints) per second. We use one second as the context window size to create the visual and path history (\ie, $R=5$), and produce one-second future scanpaths (\ie, $S=5$). As for predicting scanpaths that are longer than one second, we apply our PID controller-based sampling strategy multiple rounds, as described in Sec.~\ref{sec:sample}.

We set the quantization step size $\Delta$ in Eq.~\eqref{eq:uq} to $0.2$ (with a quantization error $<0.034\degree$). The resolution of the extracted viewport is set to $H_v \times W_v=252\times 484$, covering an FoV of $\phi_v \times\theta_v = {63}\degree \times {112}\degree$. As shown in Fig.~\ref{fig:SPNet}, for the historical visual context, we set $H=8$, $W=14$, $C=16$, and  $C_v=128$; for the historical path context, we set $C_h=128$; for the causal patch context, we set $C_c=32$. The number of Gaussian components $K$ in Eq.~\eqref{eq:qgmm} is set to $3$. For the PID controller, we set the sampling interval $\Delta \tau$ in Eq.~\eqref{eq:nm} as the inverse of the sampling rate, \ie, $\Delta \tau = 0.2$ second. The PID controller parameters in Eq.~\eqref{eq:pida} are set using the Ziegler–Nichols method~\cite{ziegler1942optimum} to $K_p = 0.6 K_u$, $K_i=2K_u/P_u$, and $K_d=K_u P_u / 8$, respectively. For the CVPR18 dataset, we set $K_u = 20$ and $P_u=0.29$, while for the VRW23 dataset, we set $K_u = 96$ and $P_u=0.29$. 

During model training, we first initialize the convolution layers in ResNet-50 with weights pre-trained on ImageNet, and initialize other parameters by He's method~\cite{he2015delving}. We then optimize our model by minimizing Eq.~\eqref{eq:obj} using Adam~\cite{kingma2014adam} with an initial learning rate of $10^{-4}$ and a minibatch size of $B=48$ (in parallel on $4$ NVIDIA A100 cards). We decay the learning rate by a factor of $10$ whenever training plateaus. We train two separate models, one for the CVPR18 dataset by following the training/test set splitting in~\cite{xu2018gaze} and the other for the VRW23 dataset, in which we use the first $400$ videos for training and the rest $102$ videos for testing.

We evaluate panoramic scanpath predictors from three perspectives: 1) \textit{prediction accuracy}, 2) \textit{perceptual realism}, and 3) \textit {generalization} to unseen datasets. For prediction accuracy evaluation, we introduce four quantitative metrics: the minimum orthodromic distance, minimum Levenshtein distance, minimum dynamic time warping distance, and maximum temporal correlation. Specifically, given a panoramic video, we define the set of human scanpaths, $\mathcal{S} = \{\bm s^{(i)}\}_{i=1}^{\vert\mathcal{S}\vert}$, as the ground-truths. The minimum orthodromic distance between $\mathcal{S}$ and the set of predicted $\hat{\mathcal{S}}=\{\hat{\bm s}^{(i)}\}_{i=1}^{\vert
\hat{\mathcal{S}}\vert}$ can be computed by 
\begin{align}\label{minADE}
\mathrm{minOD}\left(\mathcal{S},\hat{\mathcal{S}}\right)=\min\limits_{\bm s\in \mathcal{S}, \hat{\bm s}\in \hat{\mathcal{S}}}\mathrm{OD}\left(\bm s,\hat{\bm s}\right),
\end{align}
where the orthodromic distance\footnote{Orthodromic distance is also known as great-circle or spherical distance.} $\mathrm{OD}(\cdot,\cdot)$ between two scanpaths of the same length is defined as 
\begin{align}\label{eq:orthdis}
\begin{split}
    \mathrm{OD}(\bm s,\hat{\bm s}) = \frac{1}{T}\sum_{t=0}^{T-1}&\mathrm{arccos}\Big(\mathrm{cos}(\phi_{t})\mathrm{cos}(\hat{\phi}_{t})\mathrm{cos}(\theta_t - \hat{\theta}_t)
    \\&+\mathrm{sin}(\phi_{t})\mathrm{sin}(\hat{\phi}_{t})\Big).
\end{split}
\end{align}
Similarly, we compute the minimum Levenshtein distance ($\mathrm{minLEV}$) and minimum dynamic time warping distance ($\mathrm{minDTW}$) by replacing the orthodromic distance with Levenshtein similarity~\cite{privitera2000algorithms} and dynamic time warping terms~\cite{berndt1994using}, respectively. The maximum temporal correlation between $\mathcal{S}$ and $\hat{\mathcal{S}}$ is calculated by 
\begin{align}\label{maxTC}
   \mathrm{maxTC}\left(\mathcal{S}, \hat{\mathcal{S}}\right)=\max\limits_{\bm s\in \mathcal{S}, \hat{\bm s}\in \hat{\mathcal{S}}}\mathrm{TC}(\bm{s},\hat{\bm s}).
\end{align}
It is noteworthy that we intentionally opt for best-case set-to-set distance metrics to avoid specifying, for each predicted scanpath from $\hat{\mathcal{S}}$, one ground-truth from $\mathcal{S}$. Unlike path-to-path distances, these measures do not penalize generation diversity when measuring prediction accuracy.

\begin{table}
\scriptsize
\centering
\caption{Comparison in terms of $\mathrm{minOD}$, $\mathrm{maxTC}$, $\mathrm{minLEV}$, and $\mathrm{minDTW}$ on CVPR18~\cite{xu2018gaze}. The top two results are highlighted in bold}
\label{tab:res_cvpr}
\begin{tabular}{l@{\hspace{6pt}}c@{\hspace{6pt}}c@{\hspace{6pt}}c@{\hspace{6pt}}c}
\toprule
Model& $\mathrm{minOD} \downarrow$  & $\mathrm{maxTC} \uparrow$ &$\mathrm{minLEV}\downarrow$ & $\mathrm{minDTW}\downarrow$ \\
\midrule
Random &$1.489$&$0.161$&$284.64$&$240.45$\\
\midrule
Path-Only & $0.629$ & $0.382$ & $184.19$ & $101.95$ \\
Nguyen18 (CB-sal) & $0.779$ & $0.293$ & $201.88$ & $122.94$ \\
Nguyen18 (GT-sal) & $0.808$ & $0.277$ & $203.17$ & $124.58$ \\
Xu18 (CB-sal) & $0.977$ & $0.395$ & $195.32$ & $117.21$ \\
Xu18 (GT-sal) & $\textbf{0.522}$ & $0.467$ & $159.55$ & $74.27$ \\
TRACK (CB-sal) & $0.852$ & $0.392$ & $190.16$ & $109.61$ \\
TRACK (GT-sal) & $\textbf{0.456}$ & $0.498$ & $\mathbf{151.25}$ & $\mathbf{60.58}$ \\
\midrule
Ours-5 & $0.773$ & $\textbf{0.644}$ & $182.71$ & $88.53$ \\
Ours-20 & $0.627$ & $\textbf{0.708}$ & $\mathbf{158.92}$ & $\mathbf{68.88}$ \\
\bottomrule
\end{tabular}
\end{table}

\begin{table}
\scriptsize
\centering
\caption{Comparison in terms of $\mathrm{minOD}$, $\mathrm{maxTC}$, $\mathrm{minLEV}$, and $\mathrm{minDTW}$ on VRW23~\cite{fang2022subjective}}
\label{tab:res_our}
\begin{tabular}{l@{\hspace{6pt}}c@{\hspace{6pt}}c@{\hspace{6pt}}c@{\hspace{6pt}}c}
\toprule
Model& $\mathrm{minOD} \downarrow$ & $\mathrm{maxTC} \uparrow$ & $\mathrm{minLEV}\downarrow$ & $\mathrm{minDTW}\downarrow$ \\
\midrule
 Random & $1.571$ &$0.225$&$119.50$&$100.40$\\
\midrule
 Path-Only & $1.072$ & $0.676$ & $82.27$ & $49.28$\\
 Nguyen18 (CB-sal) & $1.141$ & $0.425$ & $107.14$ & $75.93$ \\
 Nguyen18 (GT-sal) & $1.063$ & $0.415$ & $105.41$ &  $73.30$ \\
 Xu18 (CB-sal) & $1.185$ & $0.637$ & $94.11$ & $53.03$\\
 Xu18 (GT-sal) & $1.215$ & $0.618$ & $89.01$ & $46.98$\\
 TRACK (CB-sal) & $1.067$ & $0.699$ & $87.74$ & $47.14$\\
 TRACK (GT-sal) & $0.966$ & $0.686$ & $\mathbf{74.73}$ & $34.80$\\
\midrule
 Ours-5 & $\textbf{0.645}$ & $\textbf{0.738}$ & $75.45$ & $\mathbf{30.37}$\\
Ours-20 & $\textbf{0.542}$ & $\textbf{0.796}$ & $\mathbf{68.90}$ & $\mathbf{24.87}$\\
\bottomrule
\end{tabular}
\end{table}

\begin{table*}
 \centering 
 \scriptsize
 \caption{Comparison in terms of $\mathrm{SminOD}$ and $\mathrm{SmaxTC}$ on CVPR18~\cite{xu2018gaze} and VRW23~\cite{fang2022subjective}. The slice length, $T_s$, is set to $\{5, 15\}$, corresponding to one-second and three-second sliced scanpaths, respectively. The prediction horizon, $S$, is set to the entire duration of each test video, excluding the initial frame that serves as the historical context}
 \label{tab:res_cvpr_vrw_merge}
 \begin{tabular}{l c c c c c c c c}
     \toprule
     \multirow{2}{*}{Model} 
     & \multicolumn{4}{c}{CVPR18} 
     & \multicolumn{4}{c}{VRW23} \\
     \cmidrule(r){2-5} \cmidrule(l){6-9}
     & $\mathrm{SminOD}$-$5\downarrow$ & $\mathrm{SminOD}$-$15\downarrow$ & $\mathrm{SmaxTC}$-$5\uparrow$ & $\mathrm{SmaxTC}$-$15\uparrow$ & 
     $\mathrm{SminOD}$-$5\downarrow$ & $\mathrm{SminOD}$-$15\downarrow$ & $\mathrm{SmaxTC}$-$5\uparrow$ & $\mathrm{SmaxTC}$-$15\uparrow$ \\
     \midrule
 Random
 & $1.154$ & $1.344$ & $0.756$ & $0.390$ &  $1.204$ & $1.344$ & $0.679$ & $0.390$  \\
 \midrule
 Path-Only
 & $0.282$ & $0.367$ & $0.854$ & $0.707$ &  $0.309$ & $0.386$ & $0.949$ & $0.865$  \\
 Nguyen18 (CB-sal)
 & $0.418$ & $0.508$ & $0.800$ & $0.563$ &  $0.770$ & $0.923$ & $0.718$ & $0.527$  \\
 Nguyen18 (GT-sal)
 & $0.466$ & $0.566$ & $0.814$ & $0.604$ &  $0.726$ & $0.851$ & $0.709$ & $0.523$  \\
 Xu18 (CB-sal)
 & $0.626$ & $0.724$ & $0.776$ & $0.509$ &  $0.437$ & $0.527$ & $0.795$ & $0.709$  \\
 Xu18 (GT-sal)
 & $0.236$ & $0.309$ & $0.894$ & $0.754$ &  $0.397$ & $0.511$ & $0.773$ & $0.728$ \\
 TRACK (CB-sal)
 & $0.407$ & $0.519$ & $0.931$ & $0.805$ &  $0.348$ & $0.430$ & $0.953$ & $0.878$  \\
 TRACK (GT-sal)
 & $\textbf{0.197}$ & $\textbf{0.261}$ & $0.898$ & $0.763$ &  $0.259$ & $0.335$ & $0.907$ & $0.837$  \\
 \midrule
 Ours-5
 & $0.215$ & $0.311$ & $\textbf{0.988}$ & $\textbf{0.956}$ &  $\textbf{0.171}$ & $\textbf{0.296}$ & $\textbf{0.989}$ & $\textbf{0.940}$ \\
 Ours-20
 & $\textbf{0.119}$ & $\textbf{0.190}$ & $\textbf{0.993}$ & $\textbf{0.971}$ &  $\textbf{0.118}$ & $\textbf{0.226}$ & $\textbf{0.995}$ & $\textbf{0.965}$ \\
 \bottomrule
 \end{tabular}
\end{table*}

Additionally, inspired by the time-delay embedding in dynamical systems~\cite{sauer1991embedology, wang2011simulating}, we introduce the sliced versions of the minimum orthodromic distance and maximum temporal correlation, respectively. We first slice each ground-truth scanpath $\bm s^{(i)} \in \mathcal{S}$, for $i \in {1,\ldots, \vert\mathcal{S}\vert}$, into $N_s$ overlapping sub-paths of length $T_s$, $\{\bm s^{(i)}_t\}_{t=1}^{N_s}$, in which the overlap between two consecutive sub-paths is set to $\lfloor T_s / 2 \rfloor$. This gives rise to $N_s$ sets of sliced scanpaths $\{\mathcal{S}_t\}_{t =1}^{N_s}$, where $\mathcal{S}_t = \{\bm s^{(i)}_t\}_{i=1}^{\vert\mathcal{S}\vert}$. Similarly, for the predicted scanpath set $\hat{\mathcal{S}}$, we create $N_s$ sets of sliced scanpaths $\{\hat{\mathcal{S}}_t\}_{t =1}^{N_s}$, where $\hat{\mathcal{S}}_t = \{\hat{\bm s}^{(j)}_t\}_{j=1}^{\vert\hat{\mathcal{S}}\vert}$, and compute the sliced minimum orthodromic distance and sliced maximum temporal correlation by
\begin{align}\label{minADEsliced}
\begin{split} \mathrm{SminOD}\left(\mathcal{S},\hat{\mathcal{S}}\right)=\frac{1}{N_s}\sum_{t=1}^{N_s}\mathrm{minOD}\left(\mathcal{S}_t,\hat{\mathcal{S}}_t\right),
\end{split}
\end{align}
and 
\begin{align}\label{maxTCsliced}
\begin{split} \mathrm{SmaxTC}\left(\mathcal{S},\hat{\mathcal{S}}\right)=\frac{1}{N_s}\sum_{t=1}^{N_s}\mathrm{maxTC}\left(\mathcal{S}_t,\hat{\mathcal{S}}_t\right),
\end{split}
\end{align}
respectively. $T_s$ is set to $\{5,15\}$, which corresponds to one-second and three-second sliced scanpaths, respectively. We 
append $T_s$ to each evaluate metric (\eg, $\mathrm{SminOD}$-$5$) to differentiate the two settings. After determining $T_s$, $N_s$ can be set accordingly. Generally, the OD family focuses more on pointwise local comparison, while the TC family emphasizes more on global covariance measurement.

\begin{figure*}[!tbp]
\scriptsize
\centering
\begin{minipage}{1.\linewidth}
\begin{minipage}{0.25\linewidth}
\includegraphics[width=1\linewidth]{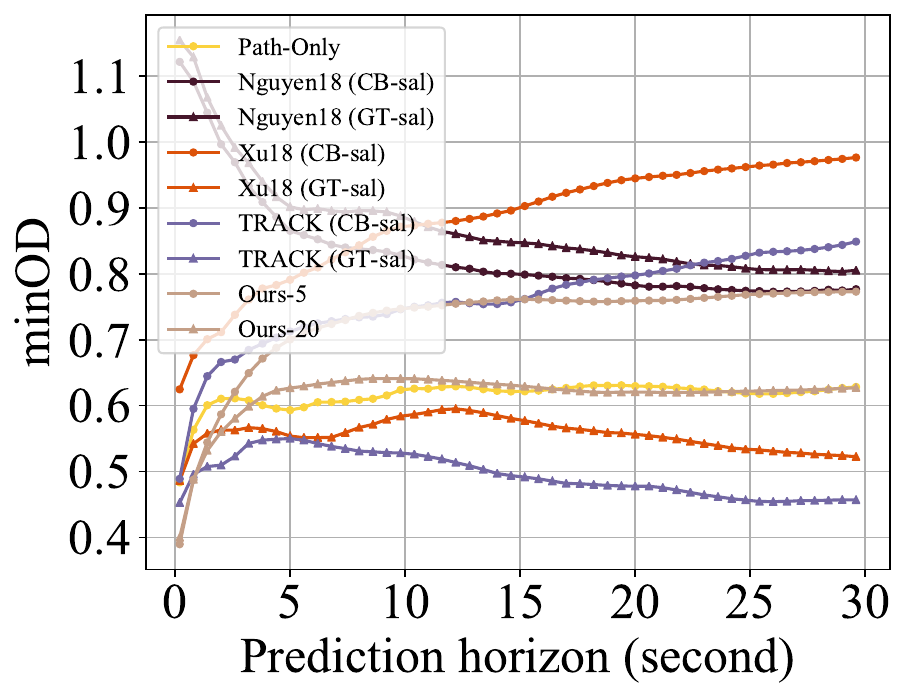}
\centering{(a)}
\end{minipage}
\begin{minipage}{0.25\linewidth}
\includegraphics[width=1\linewidth]{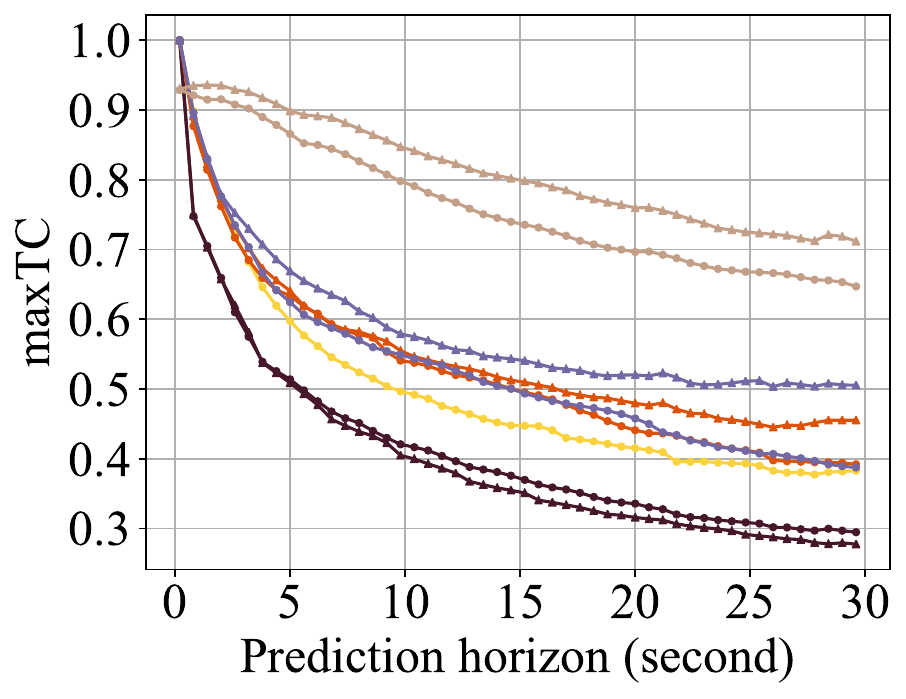}
\centering{(b)}
\end{minipage}
\begin{minipage}{0.25\linewidth}
\includegraphics[width=1\linewidth]{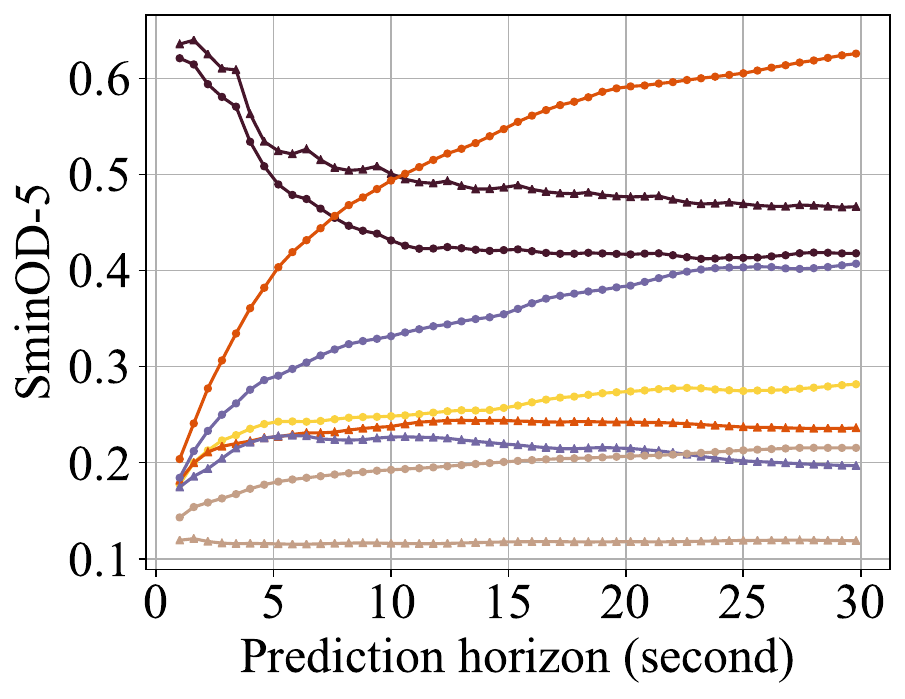}
\centering{(c)}
\end{minipage}
\begin{minipage}{0.25\linewidth}
\includegraphics[width=1\linewidth]{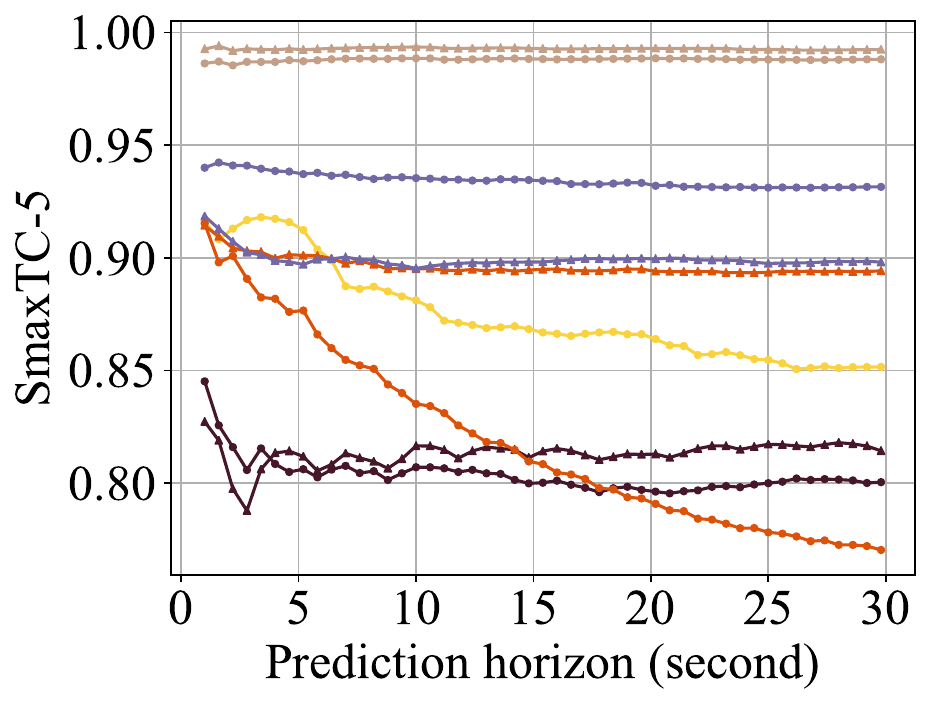}
\centering{(d)}
\end{minipage}
\end{minipage}
\caption{Performance in terms of  $\mathrm{minOD}$, $\mathrm{maxTC}$, $\mathrm{SminOD}$-$5$, and $\mathrm{SmaxTC}$-$5$ on CVPR18 as a function of prediction horizon.}\label{fig:visual_metric_cvpr}
\end{figure*}

\begin{figure*}[!tbp]
\scriptsize
\centering
\begin{minipage}{1.\linewidth}
\begin{minipage}{0.25\linewidth}
\includegraphics[width=1\linewidth]{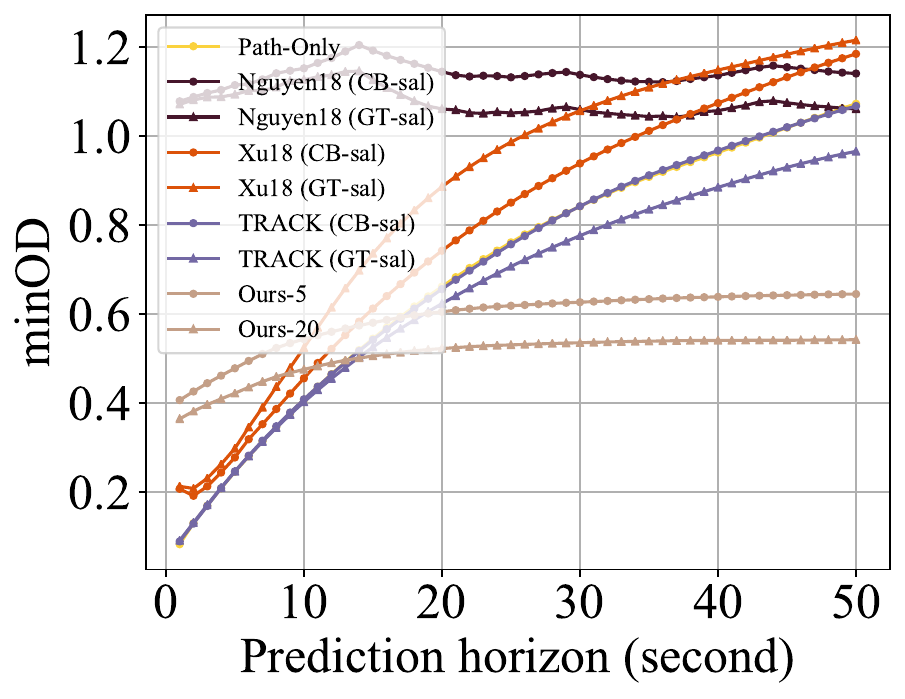}
\centering{(a)}
\end{minipage}
\begin{minipage}{0.25\linewidth}
\includegraphics[width=1\linewidth]{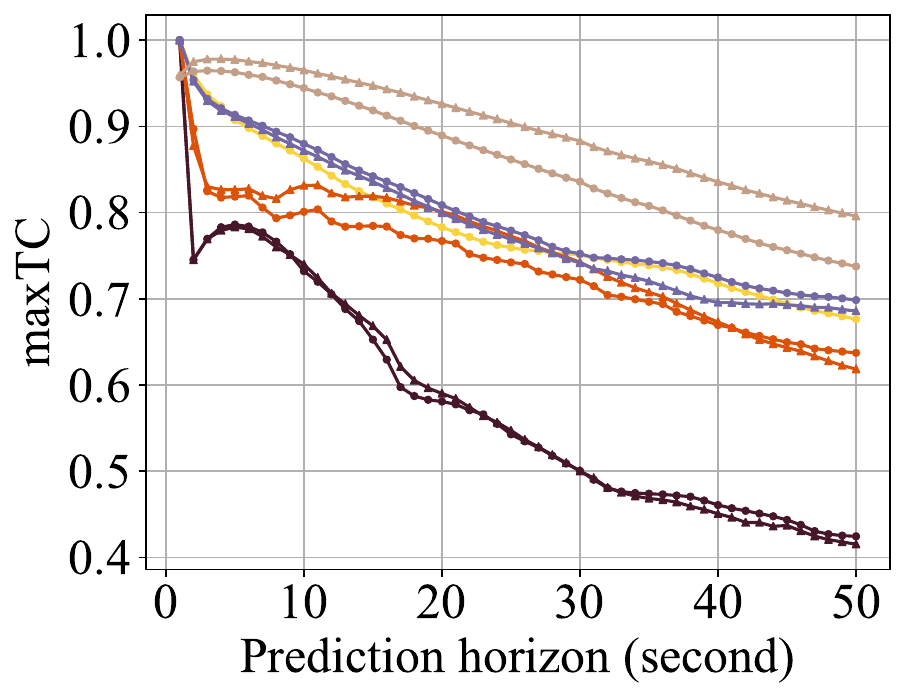}
\centering{(b)}
\end{minipage}
\begin{minipage}{0.25\linewidth}
\includegraphics[width=1\linewidth]{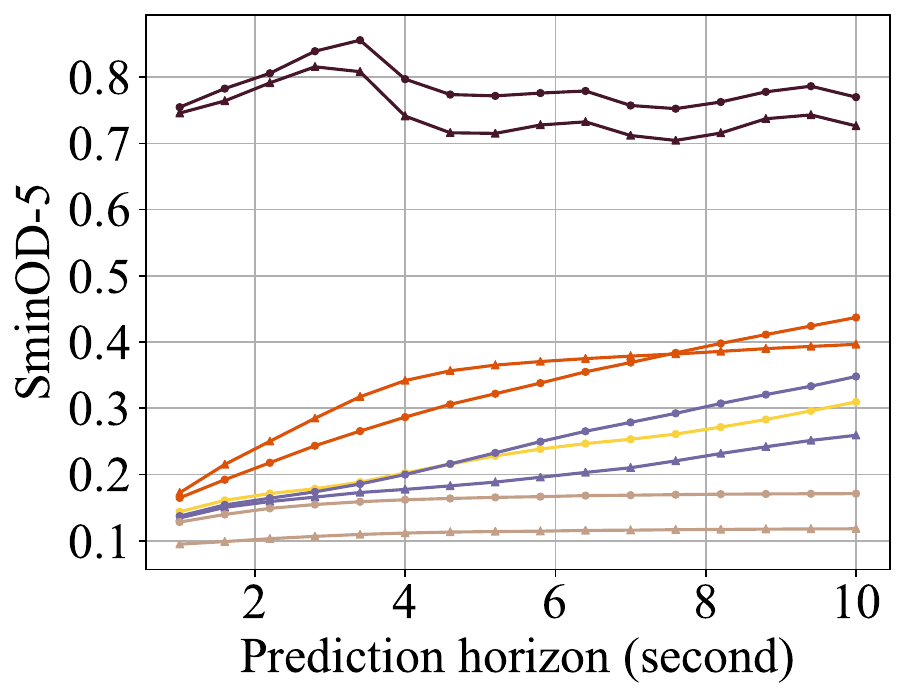}
\centering{(c)}
\end{minipage}
\begin{minipage}{0.25\linewidth}
\includegraphics[width=1\linewidth]{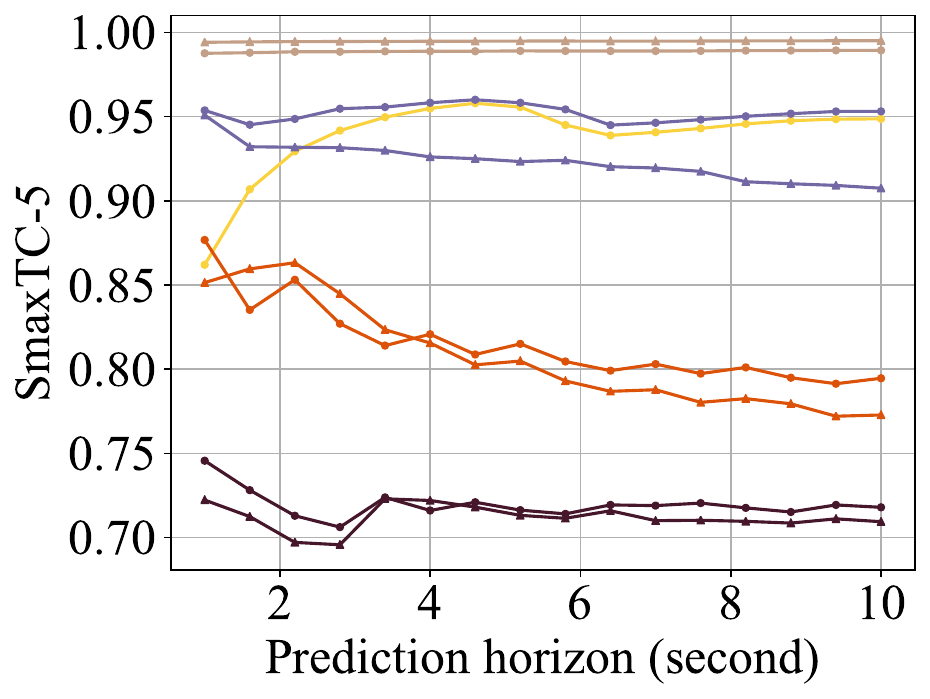}
\centering{(d)}
\end{minipage}
\end{minipage}
\caption{Performance in terms of  $\mathrm{minOD}$, $\mathrm{maxTC}$, $\mathrm{SminOD}$-$5$, and $\mathrm{SmaxTC}$-$5$ on VRW23 as a function of prediction horizon.}\label{fig:visual_metric_vrw}
\end{figure*}

For perceptual realism evaluation, we train a separate classifier for each scanpath predictor to discriminate its predicted scanpaths from those generated by humans. The underlying idea is conceptually similar to GANs~\cite{goodfellow2020generative}, except that we perform post hoc training of the classifier as the discriminator. A higher classification accuracy indicates poorer perceptual realism. Rather than solely relying on machine discrimination, we also perform a formal psychophysical experiment to quantify the perceptual realism of scanpaths (see Sec.~\ref{subsec:perrea} for implementation details).  For generalization evaluation, we test on the MMSys18~\cite{david2018dataset} and PAMI19~\cite{xu2019predicting} datasets, which consist of $19$ and $76$ distinct panoramic scenes, respectively (see Table~\ref{tab:dataset}).

\subsection{Main Experiments}\label{subsec:mainexp}

\subsubsection{Prediction Accuracy Results}\label{subsubsec:acc}

We compare the proposed method with several panoramic scanpath predictors, including a random baseline, a path-only sequence-to-sequence model~\cite{rondon2022track}, Nguyen18~\cite{nguyen2018your}, Xu18~\cite{xu2018gaze}, and TRACK~\cite{rondon2022track}.
The baseline randomly samples each viewpoint from a uniform distribution $\mathcal{U}[-\pi,\pi]$ for the longitude  and $\mathcal{U}[-\pi/2,\pi/2]$ for the latitude, respectively. This is done independently for each coordinate, initialized by the first five ground-truth viewpoints. Nguyen18, Xu18, and TRACK rely on external saliency models for scanpath prediction. We follow the experimental settings in~\cite{rondon2022track}, and exploit two types of saliency maps. The first is content-based saliency maps produced by a panoramic saliency model~\cite{nguyen2018your}, denoted by CB-sal. The second is ground-truth saliency maps aggregated spatiotemporally from multiple human viewers, denoted by GT-sal. Nevertheless, we point out two caveats when using ground-truth saliency maps. First, computing the saliency map at time $t$ may inadvertently incorporate information from future scanpaths. Second, in some competing methods (\eg, TRACK~\cite{rondon2022track}), the saliency prediction module is executed before the scanpath prediction module. Either case violates the causal assumption underlying scanpath prediction.

We re-train all competing models, following their respective training procedures. The prediction horizon $S$ for the path-only model, Nguyen18, Xu18, and TRACK during training is set to $25$, $15$, $5$, and $25$, respectively. All competing methods are deterministic, producing a single scanpath for each test panoramic video (\ie, $\vert\hat{\mathcal{S}}\vert = 1$ in Eqs.~\eqref{minADE},~\eqref{maxTC},~\eqref{minADEsliced} and~\eqref{maxTCsliced}). In stark contrast, our method is designed to be probabilistic to capture the uncertainty and diversity of scanpaths. Thus, we report the results of two variants of our method, one sampling $5$ scanpaths for each test video (\ie, $\vert
\hat{\mathcal{S}}\vert = 5$), denoted by Ours-$5$, and the other sampling $20$ scanpaths (\ie, $\vert
\hat{\mathcal{S}}\vert = 20$), denoted by Ours-$20$.

\begin{figure}[t]
\scriptsize
\centering
% \subfloat[ Video 401]{%
\subfloat{%
    \includegraphics[width=0.98\linewidth]{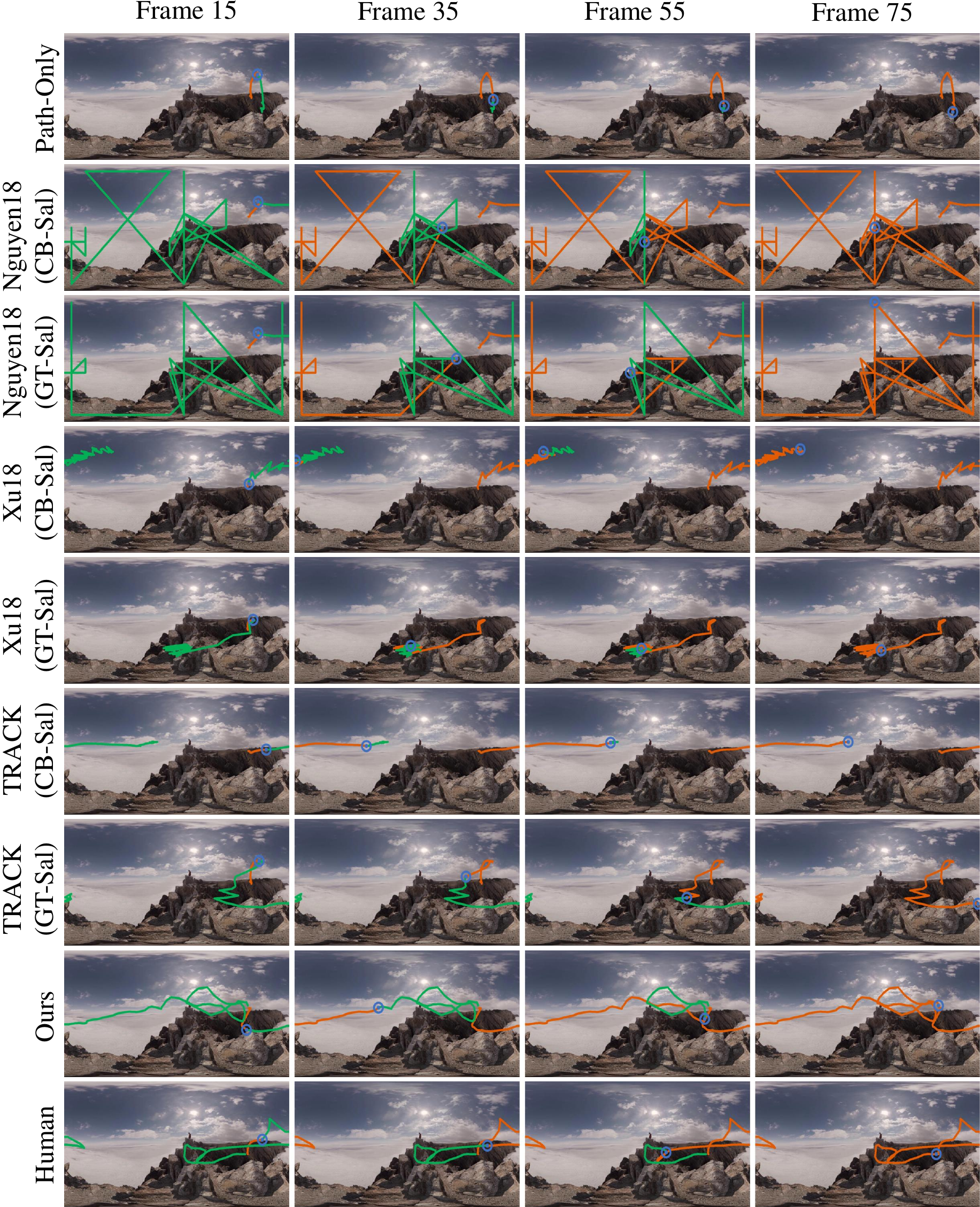}
}
% \hfill 
% \subfloat[ Video 431]{%
%     \includegraphics[width=0.48\linewidth]{fig/scanpath_v2/scanpath_3.pdf}
% }
\caption{\mk{Qualitative comparison of human and predicted scanpaths on VRW23.  In each image, the blue dot marks the current viewpoint, and the orange and green traces depict the historical and future scanpaths, respectively.}}
\label{fig:scanpath_v2}
\end{figure}

We report the $\mathrm{minOD}$, $\mathrm{maxTC}$, $\mathrm{minLEV}$, and $\mathrm{minDTW}$ results on the CVPR18 dataset in Table~\ref{tab:res_cvpr} and on the VRW23 dataset in Table~\ref{tab:res_our}, respectively. Additionally, we offer $\mathrm{SminOD}$ and $\mathrm{SmaxTC}$ results of all methods on the CVPR18 and VRW23 datasets in Table~\ref{tab:res_cvpr_vrw_merge}. The prediction horizon $S$ is set to $150$ (corresponding to 30-second scanpaths) for CVPR18 and $50$ (corresponding to 10-second scanpaths) for VRW23. From the tables, we make several interesting observations. First, the path-only model provides a highly nontrivial solution to panoramic scanpath prediction, compared with the random baseline, consistent with the observation in~\cite{rondon2022track}. This also explains the emerging but possibly ``biased'' view that the historical scanpath is all you need~\cite{chao2021transformer}. In particular, the path-only model performs better than (or at least on par with) Xu18 (CB-sal) and TRACK (CB-sal) under most metrics except for the TC family. Second, the performance of saliency-based scanpath predictors improves when the ground-truth saliency maps are allowed on CVPR18. This provides evidence that the (historical) visual context can be beneficial if it is extracted and incorporated properly. Nevertheless, such visual information may be less useful when the prediction horizon is relatively short, or even harmful with inapt incorporation, as evidenced by the TC results on VRW23. Third, our methods provide state-of-the-art performance on both datasets and under all evaluation metrics (except for $\mathrm{minOD}$ on CVPR18).

\begin{figure}[!tbp]
\scriptsize
\centering

% \subfloat[ Video 401]{%
\subfloat{%
\includegraphics[width=0.98\linewidth]{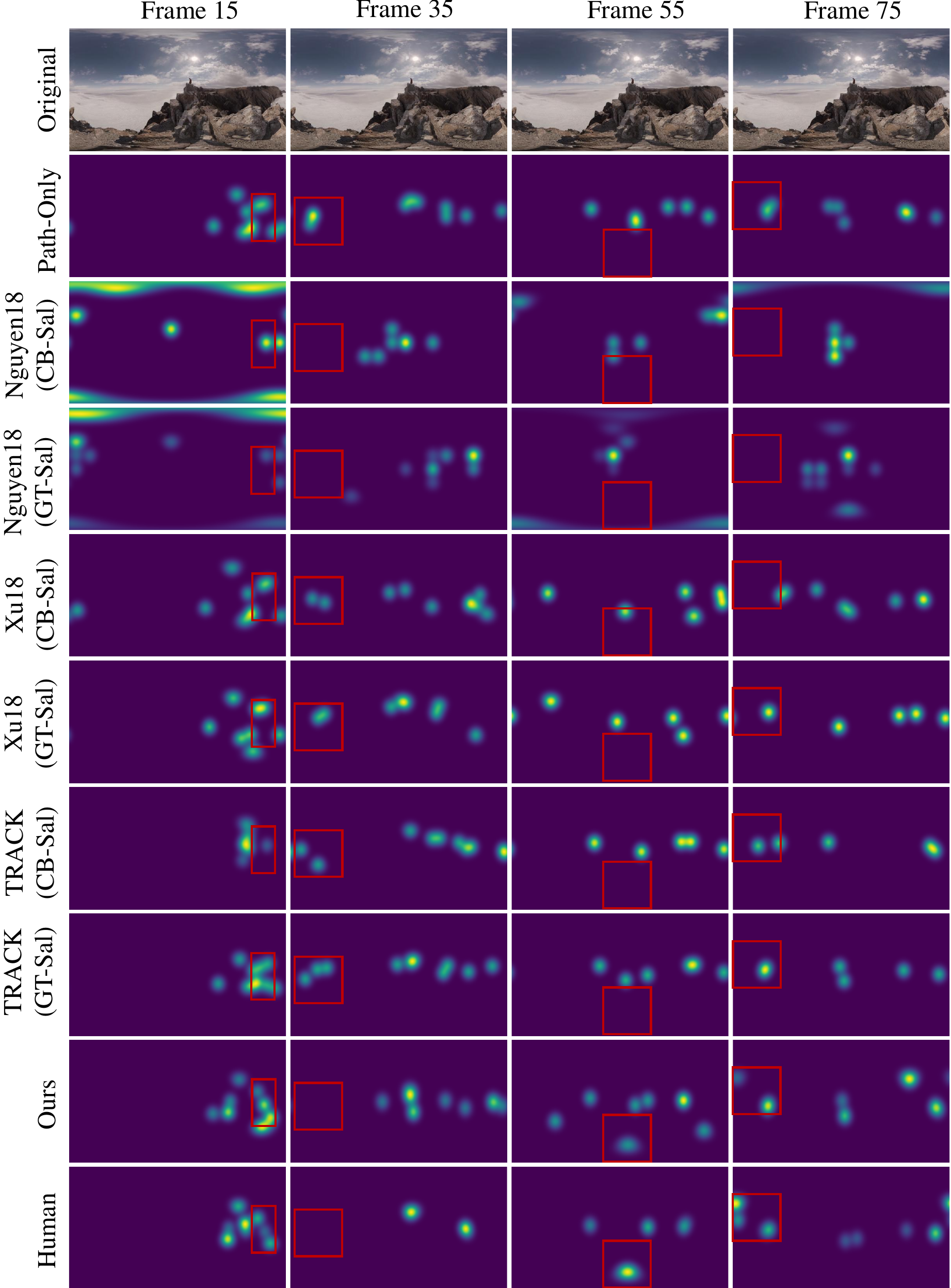}
}
% \hfill
% \subfloat[ Video 431]{%
% \includegraphics[width=0.48\linewidth]{fig/sal_v2/crop_sal_3.pdf}
% }
\caption{\mk{Qualitative comparison of saliency maps reconstructed from scanpaths on VRW23. Red boxes mark regions where our predictions align more closely with human saliency than the competing methods.}}
\label{fig:saliency_v2}
\end{figure}

\begin{table}[t]
\scriptsize
\centering
    \centering
    \caption{Comparison of saliency prediction on VRW23. NSS: normalized scanpath saliency measure. KL: Kullback–Leibler divergence}
    \label{tab:sal_vrw23}
    \begin{tabular}{l@{\hspace{6pt}}c@{\hspace{6pt}}c@{\hspace{6pt}}c@{\hspace{6pt}}c}
    \toprule
    Model& PCC $\uparrow$  & NSS~\cite{peters2005components} $\uparrow$ &AUC-Judd~\cite{judd2012benchmark} $\uparrow$ & KL $\downarrow$ \\
    \midrule
    Path-Only         & \textbf{0.749} & \textbf{2.437}& \textbf{0.942}& 1.203\\
    Nguyen18 (CB-sal) &0.471&0.863&0.619&2.482\\
    Nguyen18 (GT-sal) & 0.437&0.827 & 0.676& 2.498\\
    Xu18 (CB-sal)     & 0.643& 2.097& 0.924& 1.401 \\
    Xu18 (GT-sal)     & 0.587& 1.908& 0.917& 1.495\\
    TRACK (CB-sal)    & 0.719& 2.347& 0.940& 1.235\\
    TRACK (GT-sal)    & 0.722& 2.348& 0.940& 1.297\\
    \midrule
    Ours-20           & \textbf{0.772}& \textbf{2.483}& \textbf{0.943}& \textbf{0.749}\\
    \bottomrule
    \end{tabular}

\end{table}

 We take a closer look at the performance variations of scanpath predictors with varied prediction horizons. Figs.~\ref{fig:visual_metric_cvpr} and~\ref{fig:visual_metric_vrw} show the results under $\mathrm{minOD}$, $\mathrm{maxTC}$, $\mathrm{SminOD}$-$5$, and $\mathrm{SmaxTC}$-$5$  on CVPR18 and VRW23, respectively.  We find that initially our methods underperform slightly, but quickly catch up and significantly outperform the competing methods in the long run. This makes sense because deterministic methods are typically optimized for pointwise distance losses, and thus perform more accurately at the beginning with highly consistent viewpoints. As the prediction horizon increases, viewers tend to explore in rather different ways, leading to diverse scanpaths that cause deterministic methods to degrade. Meanwhile, we also make a ``counterintuitive'' observation: models with predicted saliency show noticeably better temporal correlation but poorer orthodromic distance than those with ground-truth saliency on VRW23 (but not on CVPR18). We posit that these discrepancies arise from the interaction between dataset characteristics (\eg, panoramic video duration) and metric emphasis (\ie, local, pointwise vs. global, listwise). Moreover, our methods remain stable when evaluated with sliced metrics.

\mk{Fig.~\ref{fig:scanpath_v2}  visually compares human and predicted scanpaths. We randomly select one human scanpath from the test video as the reference. For each competing method, we choose the prediction most similar to the reference, measured by $\mathrm{maxTC}$. Our method delivers the most accurate long-horizon predictions, closely matching the reference in both trajectory similarity and smoothness. In contrast, the path-only model shows localized confinement, while Nguyen18 and Xu18 produce jittery, erratic motions. These results indicate that our model is capable of generating realistic human-like scanpaths.}

\mk{We additionally evaluate saliency prediction on VRW23. For each predictor, we generate $20$ one-second scanpaths and aggregate all viewpoints at each timestamp to produce saliency maps~\cite{rondon2020unified}. Performance is measured using PCC, normalized scanpath saliency (NSS)~\cite{peters2005components}, AUC-Judd~\cite{judd2012benchmark}, and Kullback–Leibler (KL) divergence. As reported in Table~\ref{tab:sal_vrw23}, our model outperforms all competing predictors across all four metrics, even though most competitors take saliency as input. Fig.~\ref{fig:saliency_v2} further shows that our maps align more closely with human ground-truths, particularly in the highlighted regions.}

 \begin{figure}[!tbp]
\scriptsize
\centering
\begin{tikzpicture}
\node[inner sep=0pt] (nd0) at (0,0) {\includegraphics[width=1\linewidth]{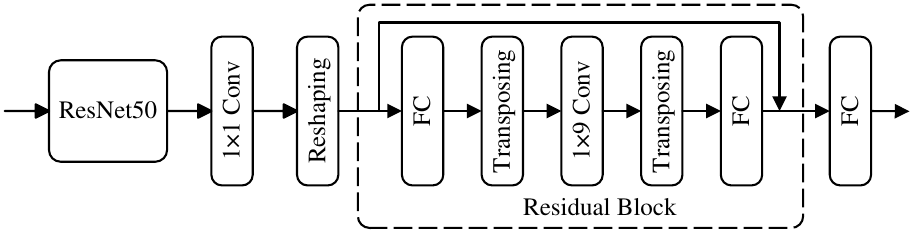}};
\node[text width=3cm,align=center] (t3) at (1.1,1.3) {\baselineskip=3pt \scriptsize{${\times 4}$} \par};
\end{tikzpicture}
\caption{Classifier specification for perceptual realism testing of scanpaths.}\label{fig:classifer}
\end{figure}

\subsubsection{Perceptual Realism Results}\label{subsec:perrea} Apart from prediction accuracy, we also evaluate the perceptual realism of predicted scanpaths. We first take a machine discrimination approach, training DNN-based binary classifiers to discriminate whether input viewport sequences are real or fake (\ie, sampled along human or model-predicted scanpaths). As shown in Fig.~\ref{fig:classifer}, we adopt a variant of ResNet-50 (the same as in Sec.~\ref{sec:visual_context}) to extract visual features from $B$ input viewport sequences with $L$ frames, leading to an intermediate representation of size $B \times L \times C \times H \times W$. We then reshape it to $(B \times L) \times (C \times H \times W)$, followed by four residual blocks and a back-end FC layer to produce an output representation of size $B \times L$. Inspired by the multi-head attention in~\cite{vaswani2017attention}, our residual block consists of a front-end FC layer, a transposing operation, a 2D convolution with a kernel size of $1\times9$, a second transposing operation, and a back-end FC layer with a skip connection. After the front-end FC layer, we split the representation into $D$ parts, with the size of $(B \times L) \times (D \times E)$, which is transposed to $B \times D\times E \times L$. We then apply 2D convolution, and transpose the convolved representation back to $(B \times L) \times (D \times E)$. We further process it with the back-end FC layer to generate the output of size $(B \times L) \times (C \times H \times W)$, which is added to the input. Last, we take the average of the output features along the time dimension, and add a sigmoid activation to estimate the real vs. fake probabilities.

\begin{table}
 \centering 
 \caption{Perceptual realism comparison through machine discrimination on CVPR18 and VRW23} \label{tab:cls} 
\scriptsize
 \begin{tabular}{l c c c c c c}
     \toprule 
     \multirow{2}{*}{Model} & \multicolumn{3}{c}{\centering{CVPR18}} & \multicolumn{3}{c}{\centering{VRW23}}\\
     \cmidrule{2-4} \cmidrule{5-7}
      & $\mathrm{Acc} \downarrow$ & $F_1 \downarrow$ & $\mathrm{BCE}\uparrow$ & $\mathrm{Acc} \downarrow$ & $F_1\downarrow$ & $\mathrm{BCE}\uparrow$ \\
     \midrule 
     Path-Only & $0.992$ & $0.992$ & $0.027$ & $0.962$& $0.962$ & $0.110$ \\
     Nguyen18 (CB-sal)& $0.999$ & $0.999$ & $0.005$& $0.996$ & $0.996$ & $0.007$\\
     Nguyen18 (GT-sal)& $0.999$ & $0.999$ & $0.002$ & $0.994$ & $0.994$ & $0.024$\\
     Xu18 (CB-sal)   & $0.980$ & $0.981$ & $0.061$& $0.978$ & $0.978$ & $0.094$ \\
     Xu18 (GT-sal)  & $0.999$ & $0.999$ & $0.008$& $0.995$ & $0.995$ & $0.022$ \\
     TRACK (CB-sal)  & $0.993$ & $0.993$ & $0.023$& $0.949$ & $0.950$ & $0.154$ \\
     TRACK (GT-sal) & $0.970$ & $0.971$ & $0.094$& $0.955$ & $0.955$ & $0.162$  \\
     \midrule
     Ours-5 & $\textbf{0.949}$ & $\textbf{0.854}$ & $\textbf{0.144}$ & $\textbf{0.868}$ & $\textbf{0.597}$ & $\textbf{0.329}$  \\
     \bottomrule
    \end{tabular}
 \end{table}

We train the classifiers by minimizing the BCE loss, with the training procedures described in Sec.~\ref{subsec:expsetup}. We test the classifiers using the classification accuracy,  $F_1$ score, as well as BCE. As shown in Table~\ref{tab:cls}, our method outperforms competing approaches on both datasets. Results are uniformly higher on VRW23, likely due to its shorter video durations. In other words, longer prediction horizons accumulate more errors, which are more readily detected by the classifiers.

\begin{table*}
 \centering 
 \scriptsize
 \caption{Comparison in terms of $\mathrm{minOD}$ and $\mathrm{maxTC}$, and their sliced versions $\mathrm{SminOD}$ and $\mathrm{SmaxTC}$ on MMSys18~\cite{david2018dataset}} \label{tab:res_cross_mm} 
 %\resizebox{\linewidth}{!}{%
 \begin{tabular}{l c c c c c c c c}
    \toprule 
    \multirow{2}{*}{Model} & \multicolumn{4}{c}{\centering{CVPR18-Trained}} & \multicolumn{4}{c}{\centering{VRW23-Trained}}\\
     \cmidrule{2-5} \cmidrule{6-9}
      & $\mathrm{minOD} \downarrow$ & $\mathrm{SminOD}$-$5 \downarrow$ & $\mathrm{maxTC} \uparrow$ & $\mathrm{SmaxTC}$-$5 \uparrow$ & $\mathrm{minOD} \downarrow$ & $\mathrm{SminOD}$-$5 \downarrow$ & $\mathrm{maxTC} \uparrow$ & $\mathrm{SmaxTC}$-$5 \uparrow$  \\
\midrule 
Path-Only 	&  $0.441$ & $0.179$  & $0.795$ & $0.914$ &  $0.577$ & $0.267$  & $0.791$ & $0.959$  \\
TRACK (CB-sal) &  $0.578$ & $0.258$  & $0.773$ & $0.967$ &  $0.617$ & $0.299$  & $0.790$& $0.971$  \\
TRACK (GT-sal) &  $0.493$ & $0.212$  & $0.714$ & $0.949$ & $0.595$ & $0.283$ & $0.729$ & $0.961$  \\
\midrule
Ours-$5$ &  $\textbf{0.416}$ & $\textbf{0.141}$  & $\textbf{0.882}$ & $\textbf{0.996}$ &  $\textbf{0.435}$ & $\textbf{0.148}$  & $\textbf{0.887}$ & $\textbf{0.997}$  \\
Ours-$20$ &  $\textbf{0.322}$ & $\textbf{0.093}$  & $\textbf{0.919}$ & $\textbf{0.998}$ &  $\textbf{0.344}$ & $\textbf{0.098}$  & $\textbf{0.923}$& $\textbf{0.998}$  \\
     \bottomrule
    \end{tabular}%
   % }
 \end{table*}

 \begin{table*}
 \centering 
 \scriptsize
 \caption{Comparison in terms of $\mathrm{minOD}$ and $\mathrm{maxTC}$, and their sliced versions $\mathrm{SminOD}$ and $\mathrm{SmaxTC}$ on PAMI19~\cite{xu2019predicting}} \label{tab:res_cross_pami} 
 %\resizebox{\linewidth}{!}{%
 \begin{tabular}{l c c c c c c c c}
    \toprule 
    \multirow{2}{*}{Model} & \multicolumn{4}{c}{\centering{CVPR18-Trained}} & \multicolumn{4}{c}{\centering{VRW23-Trained}}\\
     \cmidrule{2-5} \cmidrule{6-9}
      & $\mathrm{minOD} \downarrow$ & $\mathrm{SminOD}$-$5 \downarrow$ & $\mathrm{maxTC} \uparrow$ & $\mathrm{SmaxTC}$-$5 \uparrow$ & $\mathrm{minOD} \downarrow$ & $\mathrm{SminOD}$-$5 \downarrow$ & $\mathrm{maxTC} \uparrow$ & $\mathrm{SmaxTC}$-$5 \uparrow$  \\
\midrule 
Path-Only 	&  $\textbf{0.125}$ & $\textbf{0.064}$  & $0.636$ & $0.855$ &  $\textbf{0.593}$ & $\textbf{0.353}$  & $\textbf{0.729}$ & $0.962$  \\
TRACK (CB-sal) & $0.538$ & $0.294$  & $0.635$ & $0.951$ &  $0.986$ & $0.577$  & $0.718$ & $0.964$  \\
TRACK (GT-sal) &  $\textbf{0.174}$ & $\textbf{0.068}$  & $0.645$ & $0.922$ &  $0.646$ & $0.387$  & $0.702$ & $0.957$  \\
\midrule
Ours-$5$ &  $0.584$ & $0.408$  & $\textbf{0.801}$ & $\textbf{0.994}$ &  $0.824$ & $0.499$  & $0.624$ & $\textbf{0.996}$  \\
Ours-$20$ &  $0.346$ & $0.180$  & $\textbf{0.898}$ & $\textbf{0.999}$ &  $\textbf{0.564}$ & $\textbf{0.211}$  & $\textbf{0.747}$ & $\textbf{0.999}$  \\
     \bottomrule
    \end{tabular}%
   % }
 \end{table*}

We next take a psychophysical approach, inviting human subjects to judge whether the viewport sequences are real or not. We select $11$ and $12$ panoramic videos from the CVPR18 and VRW23 test sets, respectively. For each test video,  we generate $7$ viewport sequences by sampling along scanpaths produced by the path-only model, Xu18 (CB-sal), Xu18 (GT-sal), TRACK (CB-sal), TRACK (GT-sal), the proposed method, and one human viewer (as the real instance). 
All viewport videos are shown in the actual resolution of $252\times 448$, with a framerate of $30$ fps\footnote{We upconvert the framerate from the default $5$ fps to $30$ fps using spherical linear interpolation~\cite{shoemake1985animating}.} and in a randomized temporal order. 
Each video can be replayed multiple times until the subject is confident with her/his rating. 
We gather data from $10$ subjects for each video with normal and correct-to-normal visual acuity.  They have general knowledge of image processing, but do not know the detailed purpose of the study. We include a training session to familiarize them with the user interface and the motion patterns of real viewport sequences. 

\begin{figure}[!tbp]
\scriptsize
\centering
\includegraphics[width=0.85\linewidth]{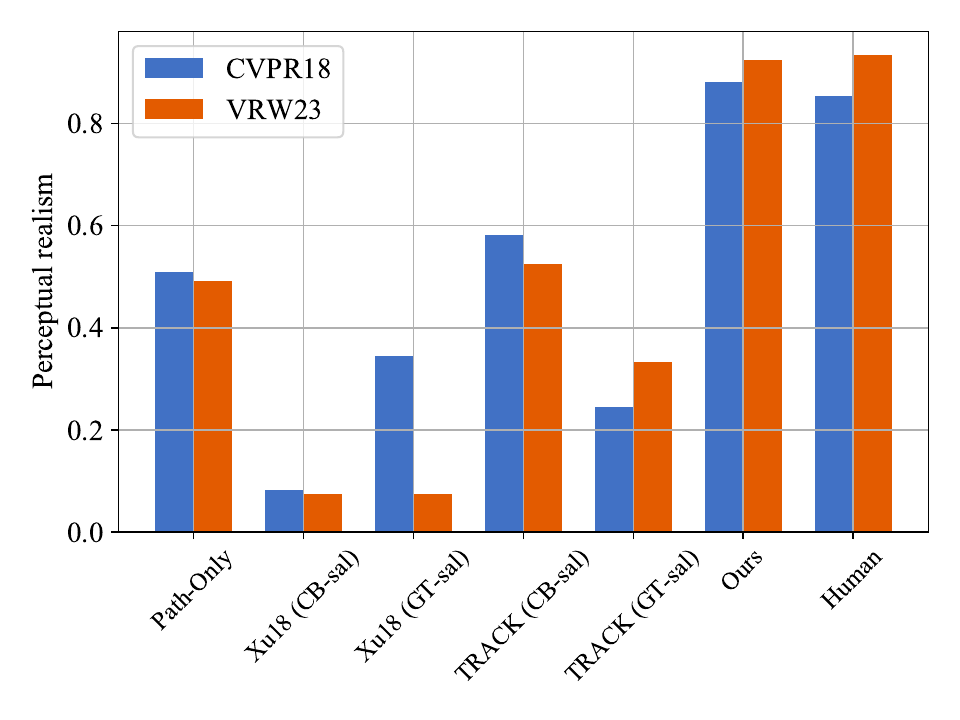}
\caption{Perceptual realism results on CVPR18 and VRW23.}
\label{fig:reality_comparison}
\end{figure}

A model's perceptual realism is the proportion of its viewport sequences labeled ``real.'' As shown in Fig.~\ref{fig:reality_comparison}, the perceptual realism of scanpaths by our model is very close to the ground-truths, and is much better than those by the competing methods on both datasets. This is due primarily to the accurate probabilistic modeling of the uncertainty and diversity of scanpaths and the PID controller-based sampler that takes into account Newton’s laws of motion. It is also interesting to note that TRACK (CB-sal) ranks third, which is consistent with the results in Fig.~\ref{fig:visual_metric_cvpr}(d) and Fig.~\ref{fig:visual_metric_vrw}(d), indicating the more perceptual relevance of the TC metric family.

\subsubsection{Cross-Dataset Generalization Results}
To test the generalizability of CVPR18-trained and VRW23-trained models, we conduct cross-dataset experiments on two relatively smaller datasets:  MMSys18~\cite{david2018dataset} and  PAMI19~\cite{xu2019predicting}. Tables~\ref{tab:res_cross_mm} and~\ref{tab:res_cross_pami} show the results, in which we omit Nguyen18 and Xu18 as they are inferior to the path-only and TRACK models. Consistent with the results in the main experiments, our methods outperform the others on both datasets in terms of temporal correlation metrics (except for Ours-$5$  trained on VRW23 and tested on PAMI19). For orthodromic distance metrics, our methods achieve the best results on MMSys18, but are worse than the path-only method on PAMI19. 
Additionally, our methods trained on CVPR18 have better performance than those trained on VRW23 when tested on PAMI19, implying closer scanpath distributions between CVPR18 and PAMI19.

\begin{table}[!tbp]
 \centering 
 \caption{Ablation of different input components in terms of $\mathrm{maxTC}$ on full-sized videos. H-Path and C-Path stand for the historical and causal path contexts, respectively} \label{tab:abl_input} 
\scriptsize
 \begin{tabular}{l c c}
     \toprule 
  Model   & CVPR18 & VRW23 \\
\midrule 
 Visual & $0.517$ & $0.628$ \\
 H-Path & $0.601$ & $0.687$ \\
 Visual + H-PATH & $0.623$ & $0.712$ \\
 Visual + H-PATH + C-PATH & $0.708$ & $0.796$ \\

     \bottomrule
    \end{tabular}
 \end{table}

\subsection{Ablation Experiments}\label{sec:ablation}

\begin{table*}[!tbp]
 \centering 
 \caption{Ablation of different training objectives with different scanpath representations in terms of $\mathrm{maxTC}$ on full-sized videos} \label{tab:abl_loss} 
\scriptsize
 \begin{tabular}{l c c c c c c}
     \toprule 
      \multirow{2}{*}{Loss} & \multicolumn{3}{c}{\centering{CVPR18}} & \multicolumn{3}{c}{\centering{VRW23}}\\
     \cmidrule{2-4} \cmidrule{5-7}
    & Spherical $(\phi,\theta)$ & 3D Euclidean $(x,y,z)$ & Relative $(u,v)$ & Spherical $(\phi,\theta)$ & 3D Euclidean $(x,y,z)$ & Relative $(u,v)$ \\
\midrule 
 MSE & $0.402$ & $0.427$ & $0.449$ & $0.566$ & $0.583$ & $0.610$   \\
 MAE & $0.453$ & $0.480$& $0.495$ & $0.638$ & $0.674$ & $0.707$  \\
 Expected Code Length & $0.656$ & $0.672$ & $0.708$ & $0.735$ & $0.748$ & $0.796$ \\
     \bottomrule
    \end{tabular}
\end{table*}

\begin{table*}[!tbp]
 \centering 
 \caption{Ablation of different samplers under models with different input components in terms of $\mathrm{maxTC}$ on one-second videos} \label{tab:abl_sampler} 
\scriptsize
 \begin{tabular}{l c c c c c c c c}
     \toprule 
     \multirow{2}{*}{Model} & \multicolumn{4}{c}{\centering{CVPR18}} & \multicolumn{4}{c}{\centering{VRW23}}\\
     \cmidrule{2-5} \cmidrule{6-9}
      & Random & Max & Beam Search & PID Controller & Random & Max & Beam Search & PID Controller \\
\midrule 
Visual 	&  $0.007$ & $0.124$  & $0.159$ & $0.551$ &  $-0.001$& $0.115$  & $0.163$& $0.515$  \\
 H-Path & $0.142$ & $0.448$ & $0.406$ & $0.782$ & $0.230$ &$0.470$ & $0.479$ & $0.794$\\
Visual + H-Path &  $0.133$ & $0.451$  & $0.418$ & $0.786$ &  $0.232$ & $0.469$  & $0.483$& $0.799$  \\
Visual + H-Path + C-Path &  $0.147$ & $0.360$  & $0.349$ & $0.844$ &  $0.245$ & $0.446$  & $0.437$& $0.825$  \\
     \bottomrule
    \end{tabular}
 \end{table*}

We conduct a series of ablation experiments to justify the design choices of our model, where we set the prediction horizon $S=5$, sample $20$ scanpaths (\ie, $\vert\hat{\mathcal{S}}\vert=20$), and report the $\mathrm{maxTC}$ results for one-second or full-sized videos.

\noindent\textbf{Input Component.} We first ablate the three input components in our model by training 1) a model with only the historical visual context, 2) a model with only the historical path context, 3) a model with the historical visual and path contexts, and 4) the full model with all three input components. We report the $\mathrm{maxTC}$ results in Table~\ref{tab:abl_input}. Our results show that adding the historical path context clearly achieves better performance, particularly on VRW23. Moreover, the causal path context also contributes substantially, validating its effectiveness as an autoregressive prior. 

\noindent\textbf{Loss Function.} We next compare different loss functions: 1) MSE, 2) mean absolute error (MAE), and 3) our expected code length. When training with MSE and MAE, we retain only one prediction head for generating viewpoints. The $\mathrm{maxTC}$ results are presented in Table~\ref{tab:abl_loss}, where we observe that the expected code length significantly outperforms  MSE and MAE.

\noindent\textbf{Scanpath Representation.} We then probe different scanpath representations: 1) spherical coordinates $(\phi,\theta)$, 2) 3D Euclidean coordinates $(x,y,z)$, and 3) relative $uv$ coordinates $(u,v)$. From Table~\ref{tab:abl_loss}, we find that our relative $uv$ representation performs the best under different loss functions.

\begin{figure}[!tbp]
\scriptsize
\centering
\includegraphics[width=0.8\linewidth]{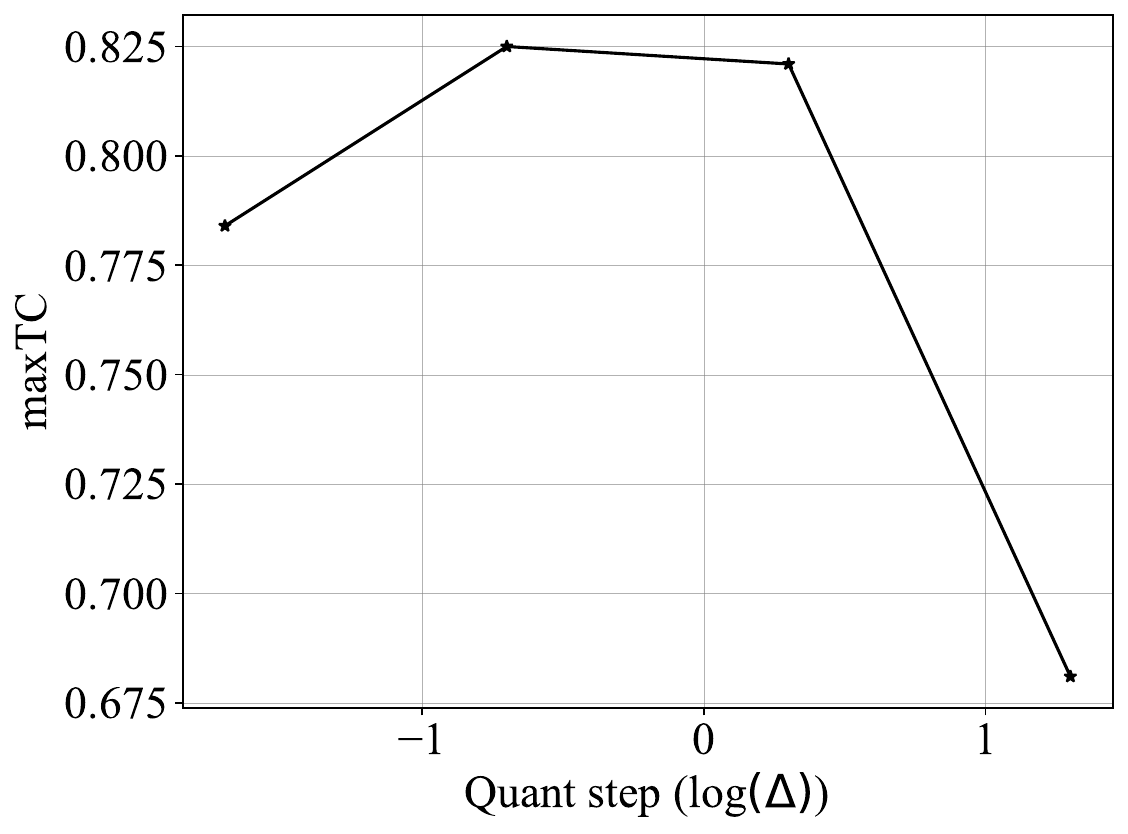}
\caption{Ablation of different quantization step sizes in terms of $\mathrm{maxTC}$.}
\label{fig:abl_quant}
\end{figure}

\noindent\textbf{Quantization Step Size.} We further study the effect of the quantization step size on probabilistic modeling, in which we test four different values of $\{0.02, 0.2, 2, 20\}$, corresponding to the largest quantization errors of $\{0.01, 0.1, 1,10\}$, respectively.  We report the $\mathrm{maxTC}$ results over one second in Fig.~\ref{fig:abl_quant}, from which we find that a proper step size is crucial for scanpath prediction. A very large step size would induce a noticeable quantization error, which impairs diversity modeling. Conversely, a very small step size would hinder the training of smooth entropy models. This provides a strong justification for the use of the discretized probability model (in Eq.~\eqref{eq:prob_acc}) over its continuous counterpart (in Eq.~\eqref{eq:qgmm}).

\noindent\textbf{Sampler.} We lastly compare our PID controller-based sampler to three counterparts: 1) a naive random sampler, 2) a max sampler, and a beam search-based sampler (with a beam width of $20$). Table~\ref{tab:abl_sampler} shows the $\mathrm{maxTC}$ results. Our PID controller-based sampler outperforms all three competing methods by a large margin.  We also observe that the causal path context increases the performance of the random sampler and our PID controller-based sampler, but decreases the performance of the max and beam search samplers. This suggests that leveraging the causal path context entails a trade-off: conditioning on inaccurate context degrades performance.

\section{Conclusion and Discussion}
We have described a new probabilistic approach to panoramic scanpath prediction from the perspective of lossy data compression. We explored a simple criterion---expected code length minimization---to train a discrete conditional probability model for quantized scanpaths. We also presented a PID controller-based sampler to generate realistic human-like scanpaths from the learned probability model.

Our method is rooted in density estimation, the mother of all unsupervised learning problems. While the question of how to reliably assess the performance of unsupervised learning methods on finite data remains open generally, we provide a quantitative measure in the context of scanpath prediction. We have carefully designed ablation experiments to point out the importance of quantization during probabilistic modeling. A similar idea that optimizes the coding rate reduction has been explored previously in image segmentation~\cite{ma2007segmentation} and recently in representation learning~\cite{dai2021closed}. 

We have advocated adopting best-case set-to-set distances to quantify similarity between human and predicted scanpaths. Our set-to-set distances can be easily generalized by first finding an optimal bipartite matching between scanpaths (for example, using the Hungarian algorithm~\cite{kuhn1955hungarian}) and then comparing pairs of matched scanpaths. We have experimented with this variant of set-to-set distances, and arrive at similar conclusions in Sec.~\ref{subsec:mainexp}.

One goal of scanpath prediction is to model and understand how humans explore different panoramic virtual scenes. Thus, we have emphasized testing the perceptual realism of predicted scanpaths via machine discrimination and human verification. Although it is relatively easy for the trained classifiers to identify predicted scanpaths, our method performs favorably in ``fooling'' human subjects, with a matched perceptual realism level to human scanpaths. Thus, our method appears promising for a number of panoramic video processing applications.

Finally, we have introduced a relative $uv$ scanpath representation in the viewport domain. It aligns with viewport sequences, simplifies computational modeling of panoramic videos, and recasts scanpath prediction as a planar problem. We believe our relative $uv$ representation has great potential in broader \mdegr computer vision tasks, including panoramic video semantic segmentation, object detection, and object tracking. In the future, we also plan to integrate audio and visual signals for panoramic scanpath prediction in the viewport domain, making our method more practically applicable.

%\nocite{*}

% \section*{Acknowledgment}
% xxxxx

%\begin{thebibliography}{1}
\bibliographystyle{IEEEtran}
\bibliography{IEEEabrv,./egbib}

% Generated by IEEEtran.bst, version: 1.14 (2015/08/26)
\begin{thebibliography}{10}
\providecommand{\url}[1]{#1}
\csname url@samestyle\endcsname
\providecommand{\newblock}{\relax}
\providecommand{\bibinfo}[2]{#2}
\providecommand{\BIBentrySTDinterwordspacing}{\spaceskip=0pt\relax}
\providecommand{\BIBentryALTinterwordstretchfactor}{4}
\providecommand{\BIBentryALTinterwordspacing}{\spaceskip=\fontdimen2\font plus
\BIBentryALTinterwordstretchfactor\fontdimen3\font minus
  \fontdimen4\font\relax}
\providecommand{\BIBforeignlanguage}[2]{{%
\expandafter\ifx\csname l@#1\endcsname\relax
\typeout{** WARNING: IEEEtran.bst: No hyphenation pattern has been}%
\typeout{** loaded for the language `#1'. Using the pattern for}%
\typeout{** the default language instead.}%
\else
\language=\csname l@#1\endcsname
\fi
#2}}
\providecommand{\BIBdecl}{\relax}
\BIBdecl

\bibitem{zhu2019prediction}
Y.~Zhu, G.~Zhai, X.~Min, and J.~Zhou, ``The prediction of saliency map for head
  and eye movements in 360 degree images,'' \emph{IEEE Transactions on
  Multimedia}, vol.~22, no.~9, pp. 2331--2344, 2019.

\bibitem{zhu2021viewing}
Y.~Zhu, G.~Zhai, Y.~Yang, H.~Duan, X.~Min, and X.~Yang, ``Viewing behavior
  supported visual saliency predictor for 360 degree videos,'' \emph{IEEE
  Transactions on Circuits and Systems for Video Technology}, vol.~32, no.~7,
  pp. 4188--4201, 2021.

\bibitem{noton1971scanpaths}
D.~Noton and L.~Stark, ``Scanpaths in saccadic eye movements while viewing and
  recognizing patterns,'' \emph{Vision Research}, vol.~11, no.~9, pp. 929--942,
  1971.

\bibitem{noton1971scanpathsineye}
------, ``Scanpaths in eye movements during pattern perception,''
  \emph{Science}, vol. 171, no. 3968, pp. 308--311, 1971.

\bibitem{perazzi2015panoramic}
F.~Perazzi, A.~Sorkine-Hornung, H.~Zimmer, P.~Kaufmann, O.~Wang, S.~Watson, and
  M.~Gross, ``Panoramic video from unstructured camera arrays,'' \emph{Computer
  Graphics Forum}, vol.~34, no.~2, pp. 57--68, 2015.

\bibitem{zoric2013panoramic}
G.~Zoric, L.~Barkhuus, A.~Engstr{\"o}m, and E.~{\"O}nnevall, ``Panoramic video:
  {Design} challenges and implications for content interaction,'' in
  \emph{European Conference on Interactive TV and Video}, 2013, pp. 153--162.

\bibitem{ng2005data}
K.-T. Ng, S.-C. Chan, and H.-Y. Shum, ``Data compression and transmission
  aspects of panoramic videos,'' \emph{IEEE Transactions on Circuits and
  Systems for Video Technology}, vol.~15, no.~1, pp. 82--95, 2005.

\bibitem{cai2022overview}
Y.~Cai, X.~Li, Y.~Wang, and R.~Wang, ``An overview of panoramic video
  projection schemes in the {IEEE} 1857.9 standard for immersive visual content
  coding,'' \emph{IEEE Transactions on Circuits and Systems for Video
  Technology}, vol.~32, no.~9, pp. 6400--6413, 2022.

\bibitem{xu2017subjective}
M.~Xu, C.~Li, Y.~Liu, X.~Deng, and J.~Lu, ``A subjective visual quality
  assessment method of panoramic videos,'' in \emph{IEEE International
  Conference on Multimedia and Expo}, 2017, pp. 517--522.

\bibitem{sitzmann2018saliency}
V.~Sitzmann, A.~Serrano, A.~Pavel, M.~Agrawala, D.~Gutierrez, B.~Masia, and
  G.~Wetzstein, ``Saliency in {VR}: How do people explore virtual
  environments?'' \emph{IEEE Transactions on Visualization and Computer
  Graphics}, vol.~24, no.~4, pp. 1633--1642, 2018.

\bibitem{rhee2017mr360}
T.~Rhee, L.~Petikam, B.~Allen, and A.~Chalmers, ``{MR360}: {Mixed} reality
  rendering for \mdegr panoramic videos,'' \emph{IEEE Transactions on
  Visualization and Computer Graphics}, vol.~23, no.~4, pp. 1379--1388, 2017.

\bibitem{lee2017high}
W.-T. Lee, H.-I. Chen, M.-S. Chen, I.-C. Shen, and B.-Y. Chen,
  ``High-resolution 360 video foveated stitching for real-time {VR},''
  \emph{Computer Graphics Forum}, vol.~36, no.~7, pp. 115--123, 2017.

\bibitem{rondon2022track}
M.~F.~R. Rondón, L.~Sassatelli, R.~Aparicio-Pardo, and F.~Precioso, ``{TRACK}:
  {A} new method from a re-examination of deep architectures for head motion
  prediction in \mdegr videos,'' \emph{IEEE Transactions on Pattern Analysis
  and Machine Intelligence}, vol.~44, no.~9, pp. 5681--5699, 2022.

\bibitem{fan2017fixation}
C.-L. Fan, J.~Lee, W.-C. Lo, C.-Y. Huang, K.-T. Chen, and C.-H. Hsu, ``Fixation
  prediction for \mdegr video streaming in head-mounted virtual reality,'' in
  \emph{Workshop on Network and Operating Systems Support for Digital Audio and
  Video}, 2017, pp. 67--72.

\bibitem{xu2018gaze}
Y.~Xu, Y.~Dong, J.~Wu, Z.~Sun, Z.~Shi, J.~Yu, and S.~Gao, ``Gaze prediction in
  dynamic \mdegr immersive videos,'' in \emph{IEEE Conference on Computer
  Vision and Pattern Recognition}, 2018, pp. 5333--5342.

\bibitem{nguyen2018your}
A.~Nguyen, Z.~Yan, and K.~Nahrstedt, ``Your attention is unique: {Detecting}
  360-degree video saliency in head-mounted display for head movement
  prediction,'' in \emph{ACM International Conference on Multimedia}, 2018, pp.
  1190--1198.

\bibitem{xu2022spherical}
Y.~Xu, Z.~Zhang, and S.~Gao, ``Spherical {DNNs} and their applications in
  \mdegr images and videos,'' \emph{IEEE Transactions on Pattern Analysis and
  Machine Intelligence}, vol.~44, no.~10, pp. 7235--7252, 2022.

\bibitem{li2018twolayer}
Y.~Li, Y.~Xu, S.~Xie, L.~Ma, and J.~Sun, ``Two-layer {FOV} prediction model for
  viewport dependent streaming of 360-degree videos,'' in \emph{International
  Conference on Communications and Networking in China}, 2018, pp. 501--509.

\bibitem{sun2021visual}
W.~Sun, Z.~Chen, and F.~Wu, ``Visual scanpath prediction using {IOR-ROI}
  recurrent mixture density network,'' \emph{IEEE Transactions on Pattern
  Analysis and Machine Intelligence}, vol.~43, no.~6, pp. 2101--2118, 2021.

\bibitem{assens2018pathgan}
M.~Assens, X.~Giro-i Nieto, K.~McGuinness, and N.~E. O'Connor, ``{PathGAN}:
  {Visual} scanpath prediction with generative adversarial networks,'' in
  \emph{European Conference on Computer Vision Workshops}, 2018, pp. 406--422.

\bibitem{martin2022scangan360}
D.~Martin, A.~Serrano, A.~W. Bergman, G.~Wetzstein, and B.~Masia,
  ``{ScanGAN360}: {A} generative model of realistic scanpaths for \mdegr
  images,'' \emph{IEEE Transactions on Visualization and Computer Graphics},
  vol.~28, no.~5, pp. 2003--2013, 2022.

\bibitem{cover2012elements}
T.~Cover and J.~Thomas, \emph{{Elements of Information Theory}}.\hskip 1em plus
  0.5em minus 0.4em\relax Wiley, 2012.

\bibitem{baltruvsaitis2018multimodal}
T.~Baltru{\v{s}}aitis, C.~Ahuja, and L.-P. Morency, ``Multimodal machine
  learning: A survey and taxonomy,'' \emph{IEEE Transactions on Pattern
  Analysis and Machine Intelligence}, vol.~41, no.~2, pp. 423--443, 2018.

\bibitem{bellman2015adaptive}
R.~Bellman, \emph{Adaptive Control Processes: A Guided Tour}.\hskip 1em plus
  0.5em minus 0.4em\relax Princeton University Press, 2015.

\bibitem{ngo2017saccade}
T.~Ngo and B.~Manjunath, ``Saccade gaze prediction using a recurrent neural
  network,'' in \emph{IEEE International Conference on Image Processing}, 2017,
  pp. 3435--3439.

\bibitem{wloka2018active}
C.~Wloka, I.~Kotseruba, and J.~K. Tsotsos, ``Active fixation control to predict
  saccade sequences,'' in \emph{IEEE Conference on Computer Vision and Pattern
  Recognition}, 2018, pp. 3184--3193.

\bibitem{xia2019predicting}
C.~Xia, J.~Han, F.~Qi, and G.~Shi, ``Predicting human saccadic scanpaths based
  on iterative representation learning,'' \emph{IEEE Transactions on Image
  Processing}, vol.~28, no.~7, pp. 3502--3515, 2019.

\bibitem{klein1988inhibitory}
R.~Klein, ``Inhibitory tagging system facilitates visual search,''
  \emph{Nature}, vol. 334, no. 6181, pp. 430--431, 1988.

\bibitem{debelen2022scanpathnet}
R.~A.~J. de~Belen, T.~Bednarz, and A.~Sowmya, ``{ScanpathNet}: {A} recurrent
  mixture density network for scanpath prediction,'' in \emph{IEEE Conference
  on Computer Vision and Pattern Recognition Workshops}, 2022, pp. 5010--5020.

\bibitem{wolfe2021guided}
J.~M. Wolfe, ``Guided search 6.0: {An} updated model of visual search,''
  \emph{Psychonomic Bulletin \& Review}, vol.~28, no.~4, pp. 1060--1092, 2021.

\bibitem{Bruce2009Saliency}
N.~D.~B. Bruce and J.~K. Tsotsos, ``Saliency, attention, and visual search:
  {An} information theoretic approach,'' \emph{Journal of Vision}, vol.~9,
  no.~3, pp. 1--24, 2009.

\bibitem{Huang2015SALICON}
X.~Huang, C.~Shen, X.~Boix, and Q.~Zhao, ``{SALICON}: {Reducing} the semantic
  gap in saliency prediction by adapting deep neural networks,'' in \emph{IEEE
  International Conference on Computer Vision}, 2015, pp. 262--270.

\bibitem{He2017Mask}
K.~He, G.~Gkioxari, P.~Dollar, and R.~Girshick, ``Mask {R-CNN},'' in \emph{IEEE
  International Conference on Computer Vision}, 2017, pp. 2961--2969.

\bibitem{assens2017saltinet}
M.~Assens, X.~Giro-i Nieto, K.~McGuinness, and N.~E. O’Connor, ``{SaltiNet}:
  {Scan-path} prediction on 360 degree images using saliency volumes,'' in
  \emph{IEEE International Conference on Computer Vision Workshops}, 2017, pp.
  2331--2338.

\bibitem{zhu2018prediction}
Y.~Zhu, G.~Zhai, and X.~Min, ``The prediction of head and eye movement for 360
  degree images,'' \emph{Signal Processing: Image Communication}, vol.~69, pp.
  15--25, 2018.

\bibitem{Felzenszwalb2010Object}
P.~F. Felzenszwalb, R.~B. Girshick, D.~McAllester, and D.~Ramanan, ``Object
  detection with discriminatively trained part-based models,'' \emph{IEEE
  Transactions on Pattern Analysis and Machine Intelligence}, vol.~32, no.~9,
  pp. 1627--1645, 2010.

\bibitem{kerkouri2022salypath360}
M.~A. Kerkouri, M.~Tliba, A.~Chetouani, and M.~Sayeh, ``{SalyPath360}:
  {Saliency} and scanpath prediction framework for omnidirectional images,'' in
  \emph{Electronic Imaging Symposium}, 2022, pp. 168--1 -- 168--7.

\bibitem{dahou2020atsal}
Y.~Dahou, M.~Tliba, K.~McGuinness, and N.~O'Connor, ``{ATSal}: {An} attention
  based architecture for saliency prediction in \mdegr videos,'' in
  \emph{International Conference on Pattern Recognition Workshops}, 2020, pp.
  305--320.

\bibitem{cornia2016deep}
M.~Cornia, L.~Baraldi, G.~Serra, and R.~Cucchiara, ``A deep multi-level network
  for saliency prediction,'' in \emph{International Conference on Pattern
  Recognition}, 2016, pp. 3488--3493.

\bibitem{lucas1981iterative}
B.~D. Lucas and T.~Kanade, ``An iterative image registration technique with an
  application to stereo vision,'' in \emph{International Joint Conference on
  Artificial Intelligence}, 1981, pp. 674--679.

\bibitem{deabreu2017look}
A.~De~Abreu, C.~Ozcinar, and A.~Smolic, ``Look around you: {Saliency} maps for
  omnidirectional images in {VR} applications,'' in \emph{International
  Conference on Quality of Multimedia Experience}, 2017, pp. 1--6.

\bibitem{pan2016shallow}
J.~Pan, E.~Sayrol, X.~Giro-i Nieto, K.~McGuinness, and N.~E. O'Connor,
  ``Shallow and deep convolutional networks for saliency prediction,'' in
  \emph{IEEE Conference on Computer Vision and Pattern Recognition}, 2016, pp.
  598--606.

\bibitem{ilg2017flownet}
E.~Ilg, N.~Mayer, T.~Saikia, M.~Keuper, A.~Dosovitskiy, and T.~Brox,
  ``{FlowNet} 2.0: {Evolution} of optical flow estimation with deep networks,''
  in \emph{IEEE Conference on Computer Vision and Pattern Recognition}, 2017,
  pp. 2462--2470.

\bibitem{xu2019predicting}
M.~Xu, Y.~Song, J.~Wang, M.~Qiao, L.~Huo, and Z.~Wang, ``Predicting head
  movement in panoramic video: {A} deep reinforcement learning approach,''
  \emph{IEEE Transactions on Pattern Analysis and Machine Intelligence},
  vol.~41, no.~11, pp. 2693--2708, 2019.

\bibitem{li2019very}
C.~Li, W.~Zhang, Y.~Liu, and Y.~Wang, ``Very long term field of view prediction
  for 360-degree video streaming,'' in \emph{IEEE Conference on Multimedia
  Information Processing and Retrieval}, 2019, pp. 297--302.

\bibitem{Cornia2018predicting}
M.~Cornia, L.~Baraldi, G.~Serra, and R.~Cucchiara, ``Predicting human eye
  fixations via an {LSTM}-based saliency attentive model,'' \emph{IEEE
  Transactions on Image Processing}, vol.~27, no.~10, pp. 5142--5154, 2018.

\bibitem{chao2021transformer}
F.-Y. Chao, C.~Ozcinar, and A.~Smolic, ``Transformer-based long-term viewport
  prediction in \mdegr video: {Scanpath} is all you need,'' in \emph{IEEE
  International Workshop on Multimedia Signal Processing}, 2021, pp. 1--6.

\bibitem{zhu2020learning}
Y.~Zhu, G.~Zhai, X.~Min, and J.~Zhou, ``Learning a deep agent to predict head
  movement in 360-degree images,'' \emph{ACM Transactions on Multimedia
  Computing, Communications, and Applications}, vol.~16, no.~4, pp. 1--23,
  2020.

\bibitem{muller2007dynamic}
M.~M{\"u}ller, \emph{Information Retrieval for Music and Motion}.\hskip 1em
  plus 0.5em minus 0.4em\relax Springer Berlin Heidelberg, 2007.

\bibitem{cohen2018spherical}
T.~S. Cohen, M.~Geiger, J.~K{\"o}hler, and M.~Welling, ``Spherical {CNNs},'' in
  \emph{International Conference on Learning Representations}, 2018.

\bibitem{esteves2018learning}
C.~Esteves, C.~Allen-Blanchette, A.~Makadia, and K.~Daniilidis, ``Learning
  {SO(3)} equivariant representations with spherical {CNNs},'' in
  \emph{European Conference on Computer Vision}, 2018, pp. 52--68.

\bibitem{jiang2019spherical}
C.~Jiang, J.~Huang, K.~Kashinath, Prabhat, P.~Marcus, and M.~Niessner,
  ``Spherical {CNNs} on unstructured grids,'' in \emph{International Conference
  on Learning Representations}, 2019.

\bibitem{wu2020spherical}
C.~Wu, R.~Zhang, Z.~Wang, and L.~Sun, ``A spherical convolution approach for
  learning long term viewport prediction in 360 immersive video,'' in
  \emph{AAAI Conference on Artificial Intelligence}, 2020, pp.
  14\,003--14\,040.

\bibitem{vaswani2017attention}
A.~Vaswani, N.~Shazeer, N.~Parmar, J.~Uszkoreit, L.~Jones, A.~N. Gomez, L.~u.
  Kaiser, and I.~Polosukhin, ``Attention is all you need,'' in \emph{Advances
  in Neural Information Processing Systems}, 2017.

\bibitem{devlin2019bert}
J.~Devlin, M.-W. Chang, K.~Lee, and K.~Toutanova, ``{BERT}: {Pre-training} of
  deep bidirectional transformers for language understanding,'' in \emph{Annual
  Conference of the North American Chapter of the Association for Computational
  Linguistics: Human Language Technologies}, 2019, pp. 4171--4186.

\bibitem{simoncelli1993distributed}
E.~P. Simoncelli, ``Distributed representation and analysis of visual motion,''
  Ph.D. dissertation, Massachusetts Institute of Technology, 1993.

\bibitem{bishop2006pattern}
C.~M. Bishop and N.~M. Nasrabadi, \emph{Pattern Recognition and Machine
  Learning}.\hskip 1em plus 0.5em minus 0.4em\relax Springer, 2006.

\bibitem{balle2016end}
J.~Ball{\'e}, V.~Laparra, and E.~P. Simoncelli, ``End-to-end optimized image
  compression,'' in \emph{International Conference on Learning
  Representations}, 2016.

\bibitem{li2020efficient}
M.~Li, K.~Ma, J.~You, D.~Zhang, and W.~Zuo, ``Efficient and effective
  context-based convolutional entropy modeling for image compression,''
  \emph{IEEE Transactions on Image Processing}, vol.~29, pp. 5900--5911, 2020.

\bibitem{li2021pseudocylindrical}
M.~Li, K.~Ma, J.~Li, and D.~Zhang, ``Pseudocylindrical convolutions for learned
  omnidirectional image compression,'' \emph{arXiv preprint arXiv:2112.13227},
  2021.

\bibitem{kingma2014adam}
D.~Kingma and J.~Ba, ``{Adam}: {A} method for stochastic optimization,'' in
  \emph{International Conference for Learning Representations}, 2015.

\bibitem{sui2022perceptual}
X.~Sui, K.~Ma, Y.~Yao, and Y.~Fang, ``Perceptual quality assessment of
  omnidirectional images as moving camera videos,'' \emph{IEEE Transactions on
  Visualization and Computer Graphics}, vol.~28, no.~8, pp. 3022--3034, 2022.

\bibitem{he2016deep}
K.~He, X.~Zhang, S.~Ren, and J.~Sun, ``Deep residual learning for image
  recognition,'' in \emph{IEEE Conference on Computer Vision and Pattern
  Recognition}, 2016, pp. 770--778.

\bibitem{devroye2006nonuniform}
L.~Devroye, \emph{Handbooks in Operations Research and Management
  Science}.\hskip 1em plus 0.5em minus 0.4em\relax Elsevier, 2006.

\bibitem{ang2005pid}
K.~H. Ang, G.~Chong, and Y.~Li, ``{PID} control system analysis, design, and
  technology,'' \emph{IEEE Transactions on Control Systems Technology},
  vol.~13, no.~4, pp. 559--576, 2005.

\bibitem{bao2016shooting}
Y.~Bao, H.~Wu, T.~Zhang, A.~A. Ramli, and X.~Liu, ``Shooting a moving target:
  {Motion-prediction-based} transmission for 360-degree videos,'' in \emph{IEEE
  International Conference on Big Data}, 2016, pp. 1161--1170.

\bibitem{wu2017dataset}
C.~Wu, Z.~Tan, Z.~Wang, and S.~Yang, ``A dataset for exploring user behaviors
  in {VR} spherical video streaming,'' in \emph{ACM Multimedia Systems
  Conference}, 2017, pp. 193--198.

\bibitem{david2018dataset}
E.~J. David, J.~Guti\'{e}rrez, A.~Coutrot, M.~P. Da~Silva, and P.~L. Callet,
  ``A dataset of head and eye movements for \mdegr videos,'' in \emph{ACM
  Multimedia Systems Conference}, 2018, pp. 432--437.

\bibitem{fang2022subjective}
Y.~Fang, Y.~Yao, X.~Sui, and K.~Ma, ``Subjective quality assessment of
  user-generated \mdegr videos,'' in \emph{IEEE Conference on Virtual Reality
  and 3D User Interfaces Abstracts and Workshops}, 2023, pp. 74--83.

\bibitem{ziegler1942optimum}
J.~G. Ziegler and N.~B. Nichols, ``Optimum settings for automatic
  controllers,'' \emph{Transactions of the American Society of Mechanical
  Engineers}, vol.~64, no.~8, pp. 759--765, 1942.

\bibitem{he2015delving}
K.~He, X.~Zhang, S.~Ren, and J.~Sun, ``Delving deep into rectifiers: Surpassing
  human-level performance on {ImageNet} classification,'' in \emph{IEEE
  Conference on Computer Vision and Pattern Recognition}, 2015, pp. 1026--1034.

\bibitem{privitera2000algorithms}
C.~M. Privitera and L.~W. Stark, ``Algorithms for defining visual
  regions-of-interest: {Comparison} with eye fixations,'' \emph{IEEE
  Transactions on Pattern Analysis and Machine Intelligence}, vol.~22, no.~9,
  pp. 970--982, 2000.

\bibitem{berndt1994using}
D.~J. Berndt and J.~Clifford, ``Using dynamic time warping to find patterns in
  time series,'' in \emph{International Conference on Knowledge Discovery and
  Data Mining}, 1994, pp. 359--370.

\bibitem{sauer1991embedology}
T.~Sauer, J.~A. Yorke, and M.~Casdagli, ``Embedology,'' \emph{Journal of
  Statistical Physics}, vol.~65, no.~3, pp. 579--616, 1991.

\bibitem{wang2011simulating}
W.~Wang, C.~Chen, Y.~Wang, T.~Jiang, F.~Fang, and Y.~Yao, ``Simulating human
  saccadic scanpaths on natural images,'' in \emph{IEEE Conference on Computer
  Vision and Pattern Recognition}, 2011, pp. 441--448.

\bibitem{goodfellow2020generative}
I.~Goodfellow, J.~Pouget-Abadie, M.~Mirza, B.~Xu, D.~Warde-Farley, S.~Ozair,
  A.~Courville, and Y.~Bengio, ``Generative adversarial networks,''
  \emph{Communications of the ACM}, vol.~63, no.~11, pp. 139--144, 2020.

\bibitem{peters2005components}
R.~J. Peters, A.~Iyer, L.~Itti, and C.~Koch, ``Components of bottom-up gaze
  allocation in natural images,'' \emph{Vision Research}, vol.~45, no.~18, pp.
  2397--2416, 2005.

\bibitem{judd2012benchmark}
T.~Judd, F.~Durand, and A.~Torralba, ``A benchmark of computational models of
  saliency to predict human fixations,'' MIT Computer Science and Artificial
  Intelligence Laboratory, Tech. Rep. MIT-CSAIL-TR-2012-001, 2012.

\bibitem{rondon2020unified}
M.~F.~R. Rondón, L.~Sassatelli, R.~Aparicio-Pardo, and F.~Precioso, ``A
  unified evaluation framework for head motion prediction methods in \mdegr
  videos,'' in \emph{ACM Multimedia Systems Conference}, 2020, pp. 279--284.

\bibitem{shoemake1985animating}
K.~Shoemake, ``Animating rotation with quaternion curves,'' in
  \emph{Proceedings of the 12th Annual Conference on Computer Graphics and
  Interactive Techniques}, 1985, pp. 245--254.

\bibitem{ma2007segmentation}
Y.~Ma, H.~Derksen, W.~Hong, and J.~Wright, ``Segmentation of multivariate mixed
  data via lossy data coding and compression,'' \emph{IEEE Transactions on
  Pattern Analysis and Machine Intelligence}, vol.~29, no.~9, pp. 1546--1562,
  2007.

\bibitem{dai2021closed}
X.~Dai, S.~Tong, M.~Li, Z.~Wu, K.~H.~R. Chan, P.~Zhai, Y.~Yu, M.~Psenka,
  X.~Yuan, and H.~Y. Shum, ``{CTRL}: Closed-loop transcription to an {LDR} via
  minimaxing rate reduction,'' \emph{Entropy}, vol.~24, no.~4, p. 456, 2022.

\bibitem{kuhn1955hungarian}
H.~W. Kuhn, ``The {Hungarian} method for the assignment problem,'' \emph{Naval
  Research Logistics Quarterly}, vol.~2, no. 1-2, pp. 83--97, 1955.

\end{thebibliography}
%\end{thebibliography}
%
\begin{IEEEbiography}
[{\includegraphics[width=1in,height=1.25in,clip,keepaspectratio]{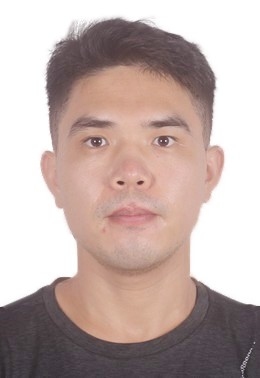}}]{Mu Li} received his B.CS in Computer Science and Technology in 2015 from Harbin Institute of Technology, and the Ph.D. degree from the Department of Computing, the Hong Kong Polytechnic University, Hong Kong, China, in 2020. He was the owner of the Hong Kong Ph.D. Fellowship. Dr. Li worked at The Chinese University of Hong Kong, Shenzhen, from 2020 to 2022. He is currently with the Harbin Institute of Technology, Shenzhen, China. His research interests include image processing, image compression, and Virtual Reality.
\end{IEEEbiography}
\begin{IEEEbiography}
[{\includegraphics[width=1in,height=1.25in,clip,keepaspectratio]{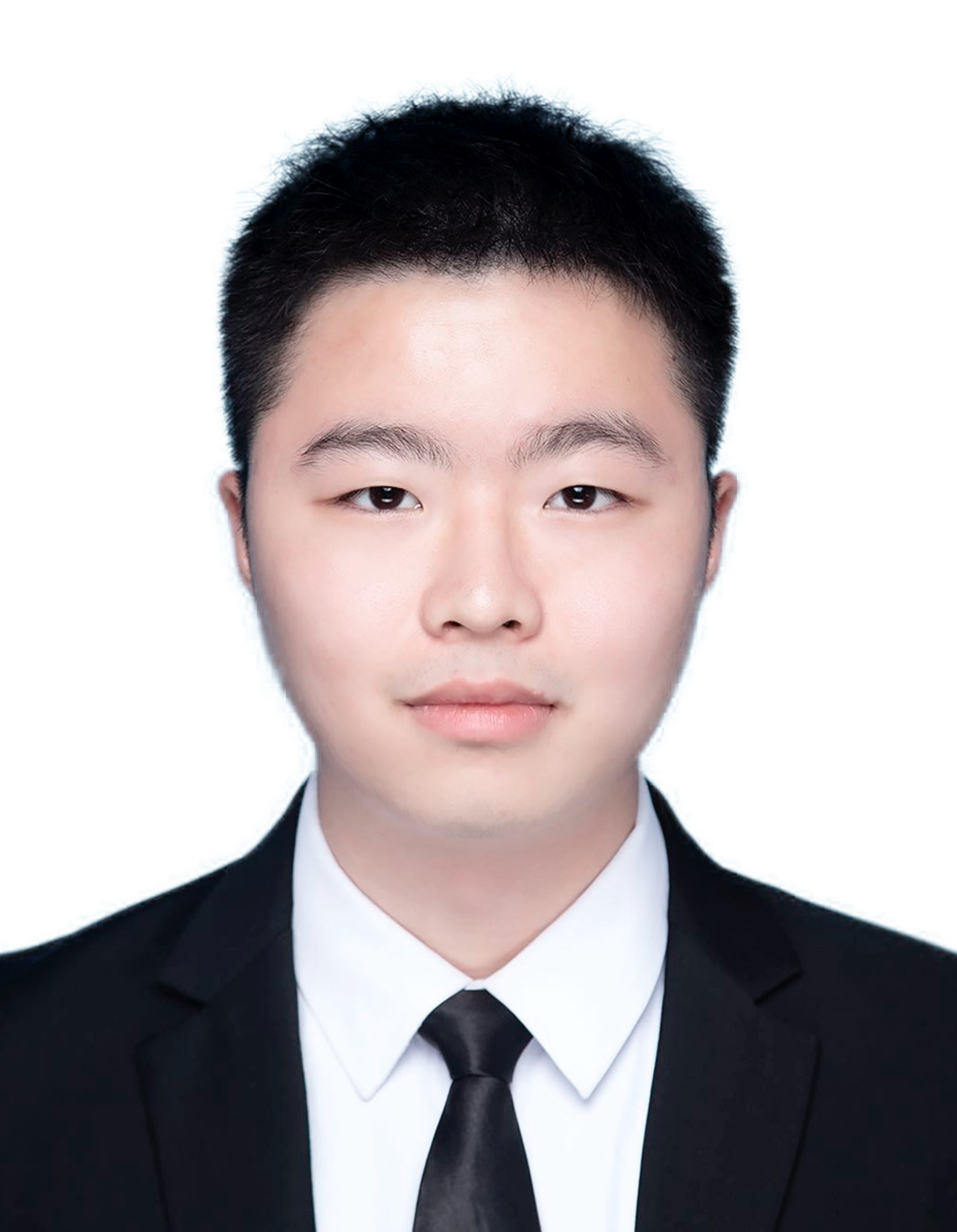}}]{Kanglong Fan} received his B.S. in Electronic Information Engineering from Beijing Institute of Technology, Beijing, China, in 2021, and the M.S. degree in Electrical Engineering from National University of Singapore, Singapore, in 2022. He is currently a Ph.D. student with City University of Hong Kong, Hong Kong, China. His research interests include Computer Vision and Virtual Reality.
\end{IEEEbiography}
\begin{IEEEbiography}[{\includegraphics[width=1in,height=1.25in,clip,keepaspectratio]{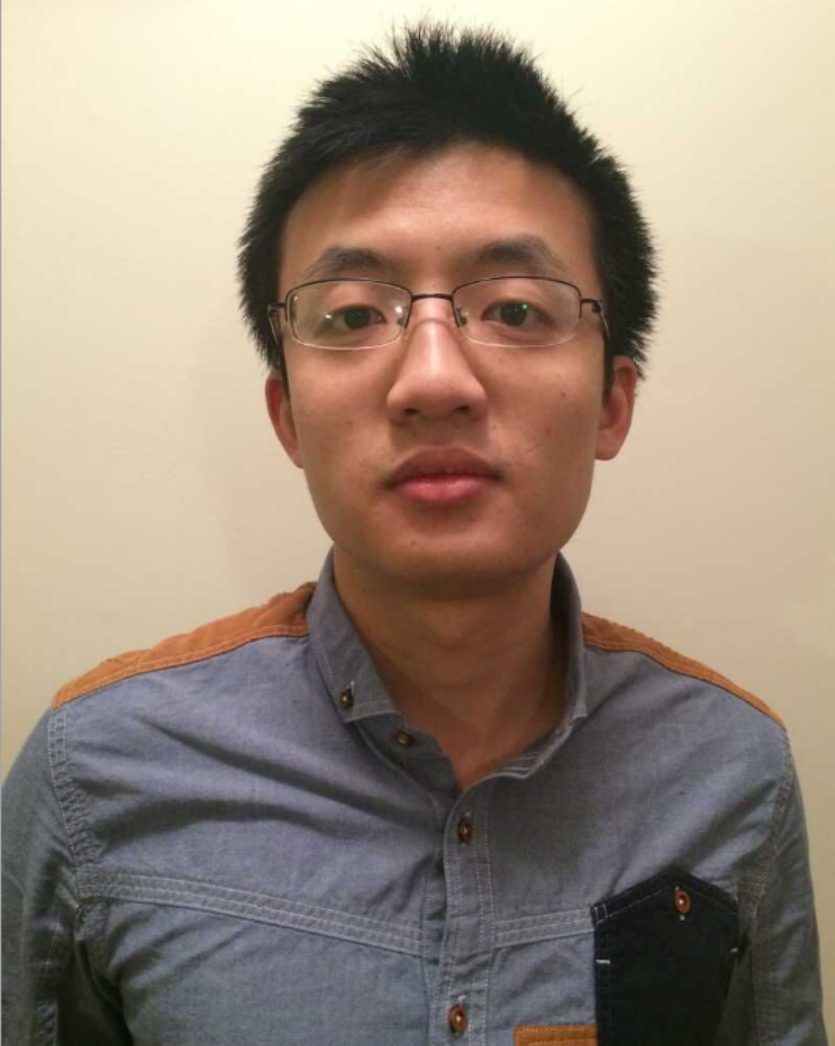}}]{Kede Ma}
(Senior Member, IEEE) received the B.E. degree from the University of Science and Technology of China (USTC) in 2012, and the MASc. and Ph.D. degrees from the University of Waterloo in 2014 and 2017, respectively. From 2018 to 2019, he was a Research Associate with the Howard Hughes Medical Institute and New York University. He is currently an Associate Professor with the Department of Computer Science, City University of Hong Kong. His research interests include computational vision, computational photography, multimedia forensics and security, and machine learning for multimedia signals. He currently serves on the Editorial Boards of IEEE Transactions on Image Processing, IEEE Transactions on Information Forensics and Security, and IEEE Signal Processing Letters. 
\end{IEEEbiography}

\end{document}